\documentclass[runningheads]{llncs}

\usepackage[final,year=2026]{eccv}

\usepackage{eccvabbrv}

\usepackage{fix-cm}
\usepackage{graphicx}
\usepackage{booktabs}
\usepackage[table]{xcolor}
\usepackage{arydshln}
\usepackage{wrapfig}
\usepackage{afterpage}
\usepackage{comment}
\usepackage{caption}
\usepackage[accsupp]{axessibility}

\usepackage[breaklinks,colorlinks,citecolor=eccvblue]{hyperref}

\usepackage{orcidlink}

\usepackage{microtype}

\usepackage{bm}

\renewcommand{\vec}[1]{\mathbf{#1}}

\newcommand{\mat}[1]{\mathbf{#1}}

\newcommand{\real}{\mathbb{R}}

\newcommand{\gmean}{\bm{\mu}}            
\newcommand{\gcovar}{\bm{\Sigma}}        

\newcommand{\gscale}{\vec{s}}            
\newcommand{\grot}{\mat{R}}              
\newcommand{\gcolor}{\vec{c}}            

\newcommand{\pose}{\bm{\theta}}           

\DeclareMathOperator{\erf}{erf}
\DeclareMathOperator{\clamp}{clamp}
\newcommand{\diag}{\operatorname{diag}}
\newcommand{\SO}{\mathrm{SO}}

\makeatletter
\renewcommand\paragraph{\@startsection{paragraph}{4}{\z@}%
  {-12\p@ \@plus -4\p@ \@minus -4\p@}%
  {-0.5em \@plus -0.22em \@minus -0.1em}%
  {\normalfont\normalsize\bfseries}}
\makeatother

\newcommand\blfootnote[1]{%
  \begingroup
  \renewcommand\thefootnote{}\footnote{#1}%
  \addtocounter{footnote}{-1}%
  \endgroup
}

\makeatletter
\let\@oldmaketitle\@maketitle%
\renewcommand{\@maketitle}{
	\@oldmaketitle%
        \begin{figure}
            \centering
            \includegraphics[width=\linewidth]{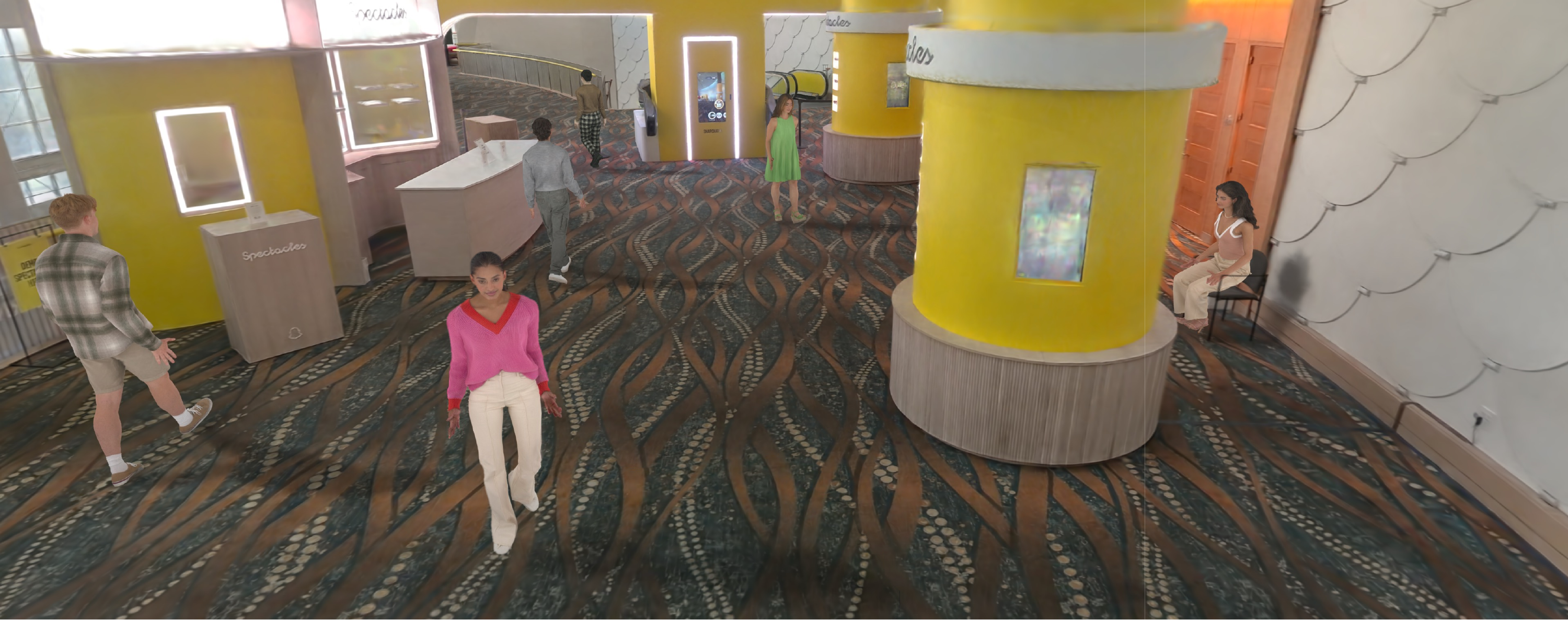}
            \caption{\textbf{RAGA casts physically plausible shadows for animated 3DGS avatars composited into 3DGS scenes.} Our ray-traced Gaussian shadow casting produces soft, temporally stable shadows entirely in Gaussian space, without any meshing.}
            \label{fig:teaser}
        \end{figure}
\vspace{-11mm}
}

\begin{document}

\title{RAGA: Real Time \texorpdfstring{\underline{Ra}}{Ra}y Traced \texorpdfstring{\underline{Ga}}{Ga}ussian Shadow Casting for 3DGS Avatar-Scene Interaction}

\titlerunning{RAGA: Ray Traced Gaussian Shadow Casting}

\author{Aymen Mir\inst{1}* \and
Riza Alp Guler\inst{3} \and
Jian Wang\inst{5} \and
Peter Wonka\inst{4, 5} \and
Bing Zhou\inst{5} \and
Gerard Pons-Moll\inst{1,2}}

\authorrunning{A.~Mir et al.}

\institute{T\"ubingen AI Center, University of T\"ubingen, Germany \and
Max Planck Institute for Informatics, Saarland Informatics Campus, Germany \and
Imperial College London, UK \and
King Abdullah University of Science and Technology (KAUST), Saudi Arabia \and
Snap Inc., USA}

\maketitle
\blfootnote{*Work partly done as intern at Snap Inc.}

\begin{abstract}
We study the problem of physically plausible shadow casting when animating 3D Gaussian Splatting (3DGS) avatars, either individually or in multi-avatar and object-interaction scenarios, within existing 3DGS scenes. In contrast to prior methods that rely on binary hit tests and mesh-based shadow casters, our method performs shadow computation entirely in Gaussian space, without requiring any mesh reconstruction. We introduce RAGA, a Ray-Traced Gaussian Shadow Casting formulation based on exact ray--Gaussian line integrals. For each occluding Gaussian, we integrate the opacity profile along the shadow ray and normalize by the theoretical maximum integral, producing a weight that captures \emph{how} the ray traverses the occluder rather than merely whether an intersection occurred. To reduce temporal variance from clothing deformations in animated avatars, we further introduce an avatar proxy representation that stabilizes shadow casting while preserving visual fidelity. We implement RAGA using custom CUDA kernels integrated with the NVIDIA OptiX framework; as such, our shadow tracer runs at rates of about 50\,FPS. We evaluate on single-avatar, multi-avatar, and avatar--object interaction scenarios across multiple datasets, demonstrating substantially improved shadow realism, temporal stability, and scene coherence. Our project page is available at \url{https://miraymen.github.io/raga/}.
\vspace{-4mm}
\end{abstract}

\section{Introduction}
Recent advances in 3D Gaussian Splatting (3DGS)~\cite{kerbl2023gaussians} have demonstrated that, beyond high-quality scene reconstruction, Gaussian-based representations can support compelling dynamic content, including animatable 3D human avatars embedded into complex real-world scenes~\cite{mir2025ahaanimatinghumanavatars, mir2025gaspacho, mir2026ahoy}. This opens the possibility of using 3DGS as a representation in content creation and simulation pipelines such as virtual production, CG compositing, and simulators for robotics and autonomous driving, which have traditionally relied on mesh-based assets.

Despite this progress, a fundamental limitation persists: physically plausible shadows are largely absent when animating 3DGS avatars interacting with static scenes or objects. As a result, inserted humans often appear visually detached from their environment, undermining realism and scene coherence, especially in dynamic settings where shadows provide critical spatial and temporal cues.

\emph{No existing method supports shadow casting at composition time (when an avatar is inserted into a new scene) in the volumetric setting where both caster (avatar) and receiver (scene) are 3D Gaussians. Tab.~\ref{tab:capability} summarizes these gaps.}

\textbf{Classical Shadow Mapping} (Tab.~\ref{tab:capability}, top block)~\cite{lokovic2000deepshadow, williams1978casting, reeves1987rendering} is not directly applicable in this setting because, unlike meshes, 3DGS is a volumetric, opacity-based representation that lacks a well-defined inside--outside structure or explicit surface geometry. Existing shadow mapping pipelines rely on surface intersections and depth tests, assumptions that do not hold for 3DGS.

\textbf{Inverse rendering methods} (Tab.~\ref{tab:capability}, middle block)~\cite{liang2023gs, gaussianshader2024, gu2025irgs, bi2024rgs, gigs2025, gsid2024, ssdgs2025, gsssr2025} decompose scenes into materials and lighting, recovering per-scene self-shadows through baked occlusion volumes~\cite{liang2023gs}, ray-traced visibility~\cite{gu2025irgs}, or learned shadow parameters~\cite{gsid2024, ssdgs2025}. However, these are per-scene optimizations on static geometry: they cannot handle dynamic avatars, and compositing a new avatar into a scene does not update the baked or learned shadow representation. 

\textbf{Ray-based splatting methods} (Tab.~\ref{tab:capability}, bottom block)~\cite{loccoz20243dgrt, wu20253dgut, Byrski2025RaySplats} perform ray--Gaussian intersection for rendering but rely on binary hit tests for shadow casting, ignoring how much of a Gaussian the ray actually traverses. Moreover, these methods assume mesh-based shadow casters; they do not address the fully volumetric setting where both caster and receiver are 3D Gaussians. Reconstructing meshes from 3DGS dynamic avatars is itself non-trivial~\cite{ye2024gaustudio} and often reduces visual fidelity.

\textbf{RAGA.} In this work, we address this gap by introducing a principled, fully Gaussian-based solution to shadow casting for animated 3DGS avatars and objects (Fig.~\ref{fig:teaser}). We propose RAGA, a Ray-Traced Gaussian Shadow Casting formulation that computes shadows using only Gaussian particles as rendering primitives, without requiring any mesh reconstruction or hybrid representation.

\definecolor{headerbg}{RGB}{200,220,240}
\definecolor{oursrow}{RGB}{220,250,220}
\begin{table}[t]
\begin{minipage}[t]{0.46\linewidth}
\vspace{0pt}\normalsize

Our method constructs a ray-traced shadow map by shooting rays from scene Gaussians toward light sources and modeling light attenuation induced by occluding Gaussians. For each occluder, we compute exact entry and exit points where the shadow ray intersects the Gaussian's support, integrate its opacity profile between those bounds, which admits a closed-form solution via the error function, and normalize by the theoretical maximum integral for that Gaussian.

\end{minipage}\hfill
\begin{minipage}[t]{0.52\linewidth}
\vspace{0pt}
\fontsize{6.5}{7.5}\selectfont
\setlength{\tabcolsep}{1pt}
\renewcommand{\arraystretch}{1.1}
\centering
\caption{\textbf{Capability Comparison}}
\label{tab:capability}
\vspace{-2mm}
\begin{tabular}{@{}lccccc@{}}
\toprule
\rowcolor{headerbg}
& \rotatebox{65}{\fontsize{5.5}{6.5}\selectfont Composable} & \rotatebox{65}{\fontsize{5.5}{6.5}\selectfont 3DGS cast.} & \rotatebox{65}{\fontsize{5.5}{6.5}\selectfont 3DGS recv.} & \rotatebox{65}{\fontsize{5.5}{6.5}\selectfont Dyn.\ 3DGS avat.} & \rotatebox{65}{\fontsize{5.5}{6.5}\selectfont Temp.\ stab.} \\
\midrule
Shadow maps~\cite{williams1978casting} & \checkmark & $\times$ & $\times$ & $\times$ & \checkmark \\
Deep shadow~\cite{lokovic2000deepshadow} & \checkmark & $\times$ & $\times$ & $\times$ & \checkmark \\
\hdashline
GS-IR~\cite{liang2023gs} & $\times$ & \checkmark & \checkmark & $\times$ & $\times$ \\
IRGS~\cite{gu2025irgs} & $\times$ & \checkmark & \checkmark & $\times$ & $\times$ \\
GS$^3$~\cite{bi2024rgs} & $\times$ & \checkmark & \checkmark & $\times$ & $\times$ \\
\hdashline
3DGRT$^\dagger$~\cite{loccoz20243dgrt} & \checkmark & $\times$ & \checkmark & $\times$ & $\times$ \\
RaySplat$^\dagger$~\cite{Byrski2025RaySplats} & \checkmark & $\times$ & \checkmark & $\times$ & $\times$ \\
\rowcolor{oursrow}
\textbf{Ours (RAGA)} & \checkmark & \checkmark & \checkmark & \checkmark & \checkmark \\
\bottomrule
\multicolumn{6}{@{}l@{}}{\fontsize{6.5}{7.5}\selectfont $^\dagger$ Can be adapted for use with 3DGS shadow casters.}
\end{tabular}
\end{minipage}
\vspace{-7mm}

\end{table}

This normalization is the key design choice: the resulting weight captures \emph{how} the ray traverses each occluder. A ray that merely grazes a Gaussian accumulates little opacity and receives a low weight; a ray passing through a long Gaussian along its major axis receives a high weight; and crucially, the same Gaussian crossed along its narrow direction receives a proportionally lower weight, correctly reflecting the reduced volumetric obstruction.

\textbf{Adaptation of ray-based methods} such as 3DGRT~\cite{loccoz20243dgrt, wu20253dgut}, or RaySplat~\cite{Byrski2025RaySplats} for shadow casting with 3D Gaussian shadow casters fails: their binary hit tests treat all intersections equally regardless of traversal depth, and 3DGRT's icosahedron proxy for each Gaussian introduces shadow artifacts due to the geometric mismatch with the smooth volumetric Gaussian. Our exact line-integral formulation eliminates these issues.

\textbf{Temporal stability.} To further reduce temporal variance caused by high-frequency clothing deformations in animated avatars, we introduce an avatar proxy representation that approximates pose-dependent changes in avatar Gaussians with isotropic Gaussians that preserve most of the geometric structure and reduce temporal variation in the avatar Gaussians. This proxy significantly reduces temporal jitter for moving avatars interacting with the environment.

We implement RAGA using custom CUDA kernels integrated with the NVIDIA OptiX framework, leveraging hardware ray tracing cores on modern GPUs; our shadow tracer alone runs at interactive rates of about 50\,FPS. We evaluate our method on multiple datasets, including single-avatar, multi-avatar, and avatar--object interaction scenarios, demonstrating substantial improvements in shadow realism, stability, and scene integration. To summarize, our contributions are as follows: \newline

\noindent1) We introduce a fully Gaussian-space shadow casting framework for 3DGS avatars animated in 3DGS scenes, without the need for mesh reconstruction.

\noindent2) We propose RAGA, a ray-traced shadow casting formulation based on exact ray--Gaussian line integrals and normalized attenuation, replacing the approximate proxy-geometry intersections used in prior work.

\noindent3)  We introduce an avatar proxy strategy that stabilizes shadows under dynamic deformations.

\noindent4) We extensively evaluate on diverse avatar and interaction scenarios, demonstrating improved realism and temporal stability over existing approaches.

\section{Related Work}

\noindent\textbf{Neural Rendering.}
NeRF~\cite{mildenhall2020nerf} drove rapid progress in neural rendering~\cite{nerf_review} but remains costly even with accelerations~\cite{mueller2022instant, barron2022mip, barron2023zipnerf, nerfstudio}. 3DGS~\cite{kerbl2023gaussians} provides an efficient explicit alternative~\cite{lassner2021pulsar, westover1991phdsplatting}, extended to dynamic scenes~\cite{shaw2023swags, luiten2023dynamic, Wu2024CVPR, lee2024ex4dgs, li2023spacetime}, SLAM~\cite{keetha2024splatam}, mesh extraction~\cite{Huang2DGS2024, guedon2023sugar}, and sparse-view NVS~\cite{mihajlovic2024SplatFields}. Existing shadow casting for Gaussians~\cite{bolanos2024gsc} does not handle shadows \emph{received} in 3DGS scenes, ray-traced variants~\cite{loccoz20243dgrt, wu20253dgut, Byrski2025RaySplats, bi2024rgs} do not cast onto scenes, and relighting methods~\cite{liang2023gs, Gao2023Relightable3DGaussians, chen2023gaussianeditor} do not address dynamic avatars. Our method casts shadows from animated 3DGS avatars onto 3DGS scenes without mesh reconstruction.

\noindent\textbf{Inverse Rendering and Shadows in 3DGS.}
Several methods decompose 3DGS scenes into materials and lighting, recovering self-shadows via baked occlusion~\cite{liang2023gs}, precomputed visibility~\cite{Gao2023Relightable3DGaussians}, environment shading~\cite{gaussianshader2024}, differentiable ray tracing~\cite{gu2025irgs}, deferred shading~\cite{gigs2025}, learned shadow parameters~\cite{gsid2024, ssdgs2025}, or screen-space ray tracing~\cite{gsssr2025}. ShadowGS~\cite{shadowgs2026} extends this to satellite imagery. All are per-scene optimizations on static geometry that cannot handle dynamic avatars or composited objects. Our method operates at composition time.

\noindent\textbf{Human Reconstruction and Rendering.}
Mesh templates~\cite{SMPL-X:2019, smpl2015loper} recover shape and pose~\cite{Bogo2016keepitsmpl, kanazawaHMR18} but lack photorealism~\cite{alldieck2018video, alldieck19cvpr, mir2020pix2surf}. Implicit representations~\cite{mescheder2019occupancy, park2019deepsdf} model detailed clothed humans~\cite{chen2021snarf, alldieck2021imghum, saito2020pifuhd, he2021arch++, huang2020arch, deng2020nasa} but struggle with re-posable rendering. Controllable NeRFs~\cite{peng2021animatable, guo2023vid2avatar, weng_humannerf_2022_cvpr, jian2022neuman, haberman2023hdhumans, heminggaberman2024trihuman, li2022tava, liu2021neural, xu2021hnerf, iqbal2023rana} and 3DGS avatars~\cite{kocabas2023hugs, qian20233dgsavatar, moreau2024human, abdal2023gaussian, zielonka2023drivable, moon2024exavatar, li2024animatablegaussians, pang2024ash, lei2023gart, xu2024gaussian, junkawitsch2025eva} produce photorealistic humans but omit scene interactions. Joint human--environment methods~\cite{xue2024hsr, tomie, mir2025gaspacho} do not address consistent shadows. Our method fills this gap.

\noindent\textbf{Humans and Scenes.}
Human--scene interaction spans affordance estimation~\cite{fouhey2014people, wang2017binge, gupta20113d}, large-scale HSI datasets~\cite{Hassan2021-gb, mir20hps, hassan2019prox, savva2016pigraphs, taheri2020grab, jiang2024scaling, cheng2023dnarendering, zhang2022couch, mahmood19amass, BABEL:CVPR:2021, Guo_2022_CVPR_humanml3d}, joint human--object reconstruction~\cite{xie2022chore, xie2023vistracker, xie2024template, zhang2020phosa}, and object-conditioned motion~\cite{zhang2022couch, starke19neural, hassan21cvpr, diller2023cghoi, zhang2022wanderings, Zhao:DartControl:2025, zhao2023dimos, mir2024origin}. These rely on mesh representations. Our method operates entirely in 3DGS space.

\noindent\textbf{Human Relighting.}
Portrait relighting uses light-stage data~\cite{Sun2019PortraitRelighting, Kanamori2018RelightingHumans, Pandey2021TotalRelighting, Ji2022GeometryAware, Kim2024SwitchLight} and diffusion models~\cite{He2024DifFRelight, Ho2020DDPM, Rombach2022LDM, Song2021SDE, Song2021DDIM}. Inverse-rendering approaches recover relightable avatars~\cite{Hasselgren2022nvdiffrecmc, Jin2023TensoIR, Guo2019Relightables, ChenLiu2022Relighting4D, Saito2024RGCA, Iwase2023RelightableHands, Chen2024URHand, Zhang2021NVPRelighting, Sarkar2023LitNeRF, WangARXIV2025}, recently extended with Gaussian-based relighting~\cite{li2024relightableandanimatable, li2024uravatar, zhan2025interactive, zhao2025sgia, fan2025rng} and subsurface scattering~\cite{dihlmann2024sssgs}. These study relighting in isolation; our method focuses on shadow casting for avatars composited into 3DGS scenes with temporal consistency.

\definecolor{figA}{RGB}{168,168,168}
\definecolor{figB}{RGB}{112,48,160}
\definecolor{figC}{RGB}{197,90,17}
\definecolor{figD}{RGB}{124,124,124}
\definecolor{figE}{RGB}{191,144,0}
\definecolor{figF}{RGB}{46,117,182}
\begin{figure*}[t]
    \centering
    \includegraphics[width=\linewidth]{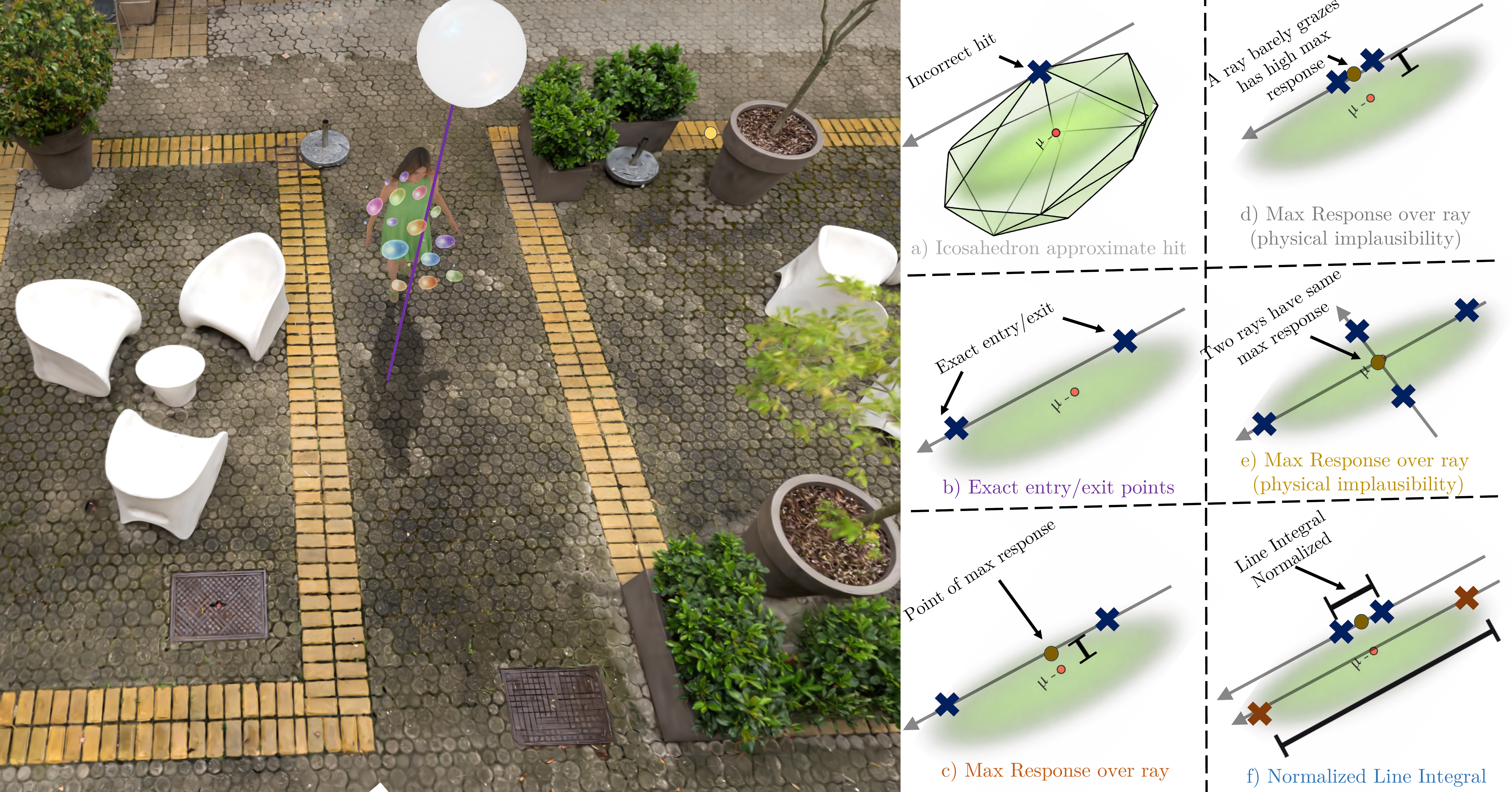}
    \caption{\textbf{Method overview.} Left: a shadow ray is cast from a scene Gaussian toward the light source through an animated 3DGS avatar by accumulating ray transmittance over all Gaussians in the avatar. Right: comparison of ray--Gaussian intersection strategies. \textcolor{figA}{(a)~The icosahedron proxy used by 3DGRT~\cite{loccoz20243dgrt} produces incorrect hits outside the Gaussian support.} \textcolor{figB}{(b)~Our exact quadratic test (Sec.~\ref{sec:intersection}) finds precise entry/exit points.} \textcolor{figC}{(c)~Evaluating the maximum Gaussian response along the ray ignores traversal depth (Sec.~\ref{sec:intersection}).} \textcolor{figD}{(d)~The point of maximum response does not reflect total volumetric obstruction (Sec.~\ref{sec:intersection}).} \textcolor{figE}{(e)~Two rays with different traversals receive the same max response, and a grazing ray is overweighted (Sec.~\ref{sec:intersection}).} \textcolor{figF}{(f)~Our normalized line integral (Sec.~\ref{sec:line-integral}, Eq.~\ref{eq:w_factor}) correctly captures how much of each Gaussian the ray traverses.}}
    \label{fig:method}
\vspace{-6mm}
\end{figure*}
\section{Method}
\label{sec:method}

Given a 3DGS scene and one or more animated 3DGS avatars, our goal is to compute physically plausible shadows cast by the avatars onto the scene at each frame, operating entirely in Gaussian space without mesh reconstruction. Fig.~\ref{fig:method} provides an overview.

Our core idea is to formulate each Gaussian's shadow contribution as a \emph{normalized line integral} along the shadow ray (Fig.~\ref{fig:method}\,\textcolor{figF}{f}). For each scene Gaussian (receiver), we cast a ray toward the light source and, for every occluding avatar Gaussian the ray encounters, we: (i)~compute exact entry and exit points where the ray intersects the Gaussian's bounded support, its $\chi^2$-confidence ellipsoid (Fig.~\ref{fig:method}\,\textcolor{figB}{b}, Sec.~\ref{sec:intersection}), rather than relying on an icosahedron proxy that can produce incorrect hits (Fig.~\ref{fig:method}\,\textcolor{figA}{a}); (ii)~integrate the Gaussian's opacity profile between those bounds, which admits a closed-form solution via the error function (Sec.~\ref{sec:line-integral}); and (iii)~normalize the result by the theoretical maximum integral, the value obtained when a ray passes through the center of the Gaussian along its longest axis (Sec.~\ref{sec:normalization}). This stands in contrast to simply evaluating the peak Gaussian response (Fig.~\ref{fig:method}\,\textcolor{figC}{c}), which is direction-agnostic and fails for anisotropic Gaussians (Fig.~\ref{fig:method}\,\textcolor{figD}{d},\,\textcolor{figE}{e}).

This normalization is the key design choice: the resulting factor $\eta\in[0,1]$ captures \emph{how} the ray traverses the Gaussian. A ray that merely grazes an occluder produces a small line integral and thus a low factor (Fig.~\ref{fig:method}\,\textcolor{figD}{d}). A ray passing through a long, thin Gaussian along its major axis accumulates substantial opacity and receives a high factor. Crucially, a ray crossing the \emph{same} Gaussian along its narrow direction receives a low factor, correctly reflecting the reduced volumetric obstruction. The factor $\eta$ then modulates the Gaussian's opacity $\alpha$ when updating the shadow transmittance (Sec.~\ref{sec:transmittance}). All ray--Gaussian queries are accelerated via a BVH built over conservative bounding boxes, enabling hardware ray tracing through NVIDIA OptiX (Sec.~\ref{sec:bvh}). Finally, to suppress temporal flickering from the pose-dependent Gaussian generator, we introduce an avatar proxy that ensures temporally coherent shadow casting (Sec.~\ref{sec:shadow_proxy}). All formal propositions and proofs supporting the derivations below are provided in the supplementary material.

\subsection{Ray--Gaussian Shadowing}
\label{sec:ray-gaussian}
We represent each occluder as an anisotropic 3D Gaussian with center $\gmean\in\real^3$, rotation $\grot\in \SO(3)$, and per-axis scales $\gscale=(s_x,s_y,s_z)$, giving the covariance
\begin{equation}
\gcovar = \grot\,\diag(s_x^2,s_y^2,s_z^2)\,\grot^{\top}.
\label{eq:cov}
\end{equation}
We use the unnormalized Gaussian kernel
\begin{equation}
g(\vec{x}) = \exp\!\Big(-\tfrac{1}{2}(\vec{x}-\gmean)^{\top}\gcovar^{-1}(\vec{x}-\gmean)\Big),
\label{eq:gauss_unnorm}
\end{equation}
whose maximum is $g(\gmean)=1$ (Lemma~1 in supp.\ mat.).

Given a receiver point $\vec{o}$ and a point light position $\bm{\ell}$, we cast a shadow ray
\begin{equation}
\vec{x}(\tau)=\vec{o}+\tau\,\vec{v},\qquad \tau\in[0,1], \quad \vec{v}=\bm{\ell}-\vec{o}.
\label{eq:ray_seg}
\end{equation}

\paragraph{Gaussian-space transform.}
To simplify the intersection test, we first define the whitening transform
\begin{equation}
\mat{L}=\diag(s_x^{-1},s_y^{-1},s_z^{-1})\,\grot^{\top},
\label{eq:L}
\end{equation}
so that $\vec{y}=\mat{L}(\vec{x}-\gmean)$ satisfies $\|\vec{y}\|_2^2 = (\vec{x}-\gmean)^{\!\top}\gcovar^{-1}(\vec{x}-\gmean)$. The ray becomes
\begin{equation}
\vec{y}(\tau)=\vec{o}' + \tau\,\vec{v}',\qquad
\vec{o}'=\mat{L}(\vec{o}-\gmean),\qquad
\vec{v}'=\mat{L}\vec{v}.
\label{eq:ray_gauss_space}
\end{equation}

\subsection{Exact Ellipsoid Intersection and Entry/Exit}
\label{sec:intersection}
We restrict contributions to the $\chi^2$-confidence ellipsoid (we use $\chi^2=9$, corresponding to a $3\sigma$ cutoff):
\begin{equation}
\|\vec{y}\|_2^2 \le \chi^2,
\label{eq:chi2_ellipsoid}
\end{equation}
which yields a quadratic constraint along the ray (Prop.~1 in supp.\ mat.):
\begin{equation}
\|\vec{o}'+\tau\vec{v}'\|_2^2
=
A \tau^2 + 2 B \tau + C,
\quad
A=\|\vec{v}'\|_2^2,\;
B=\vec{o}'^{\top}\vec{v}',\;
C=\|\vec{o}'\|_2^2.
\label{eq:quad}
\end{equation}
The maximum-response point on the ray is the minimum of Eq.~\ref{eq:quad}, attained at
\begin{equation}
\tau_c = -\frac{B}{A},
\label{eq:tcenter}
\end{equation}
and the minimum Mahalanobis distance squared to the ray is
\begin{equation}
d_{\perp}^2 = \|\vec{o}'+\tau_c\vec{v}'\|_2^2 = C - \frac{B^2}{A}.
\label{eq:dperp}
\end{equation}
If $d_{\perp}^2>\chi^2$ the ray misses the ellipsoid; otherwise we define
\begin{equation}
\Delta = \chi^2 - d_{\perp}^2 \ge 0,
\label{eq:delta}
\end{equation}
and the entry/exit times on the line are (Props.~2, Cor.~1 in supp.\ mat.)
\begin{equation}
\tau_{\mathrm{in}} = \tau_c - \sqrt{\tfrac{\Delta}{A}},
\qquad
\tau_{\mathrm{out}} = \tau_c + \sqrt{\tfrac{\Delta}{A}}.
\label{eq:tin_tout}
\end{equation}
We clamp to the ray segment,
\begin{equation}
\tilde{\tau}_{\mathrm{in}} = \max(\tau_{\mathrm{in}},0),
\qquad
\tilde{\tau}_{\mathrm{out}} = \min(\tau_{\mathrm{out}},1),
\label{eq:clamp}
\end{equation}
and discard the Gaussian if $\tilde{\tau}_{\mathrm{in}}\ge \tilde{\tau}_{\mathrm{out}}$.

\paragraph{Why not evaluate at the peak?}
A simpler alternative to integration would be to use the Gaussian's response at the closest point $\tau_c$ (Eq.~\ref{eq:tcenter}), i.e.\ $g(\vec{x}(\tau_c)) = \exp(-\tfrac{1}{2}d_{\perp}^2)$ (Fig.~\ref{fig:method}\,\textcolor{figC}{c}). However, this value depends only on the perpendicular distance $d_{\perp}$ to the Gaussian center and is entirely \emph{direction-agnostic}: two rays at the same distance but traversing vastly different distances through the Gaussian receive identical weights (Fig.~\ref{fig:method}\,\textcolor{figE}{e}). Worse, a ray barely grazing a large Gaussian along its major axis can have a small $d_{\perp}$ and thus a high peak response, despite minimal volumetric obstruction (Fig.~\ref{fig:method}\,\textcolor{figD}{d}). The line integral (Sec.~\ref{sec:line-integral}) resolves both issues by accumulating opacity over the full traversal length, naturally producing lower weights for shorter paths through the Gaussian.

\subsection{Truncated Line Integral (Closed Form)}
\label{sec:line-integral}
Along the ray, Eq.~\ref{eq:gauss_unnorm} reduces to a 1D Gaussian in $\tau$:
\begin{equation}
g(\vec{x}(\tau))
= \exp\!\Big(-\tfrac{1}{2}\|\vec{o}'+\tau\vec{v}'\|_2^2\Big)
= \exp\!\Big(-\tfrac{1}{2}d_{\perp}^2\Big)\,
\exp\!\Big(-\tfrac{1}{2}A(\tau-\tau_c)^2\Big).
\label{eq:gauss_on_ray}
\end{equation}
We integrate over the ellipsoid intersection interval,
\begin{equation}
I_{\mathrm{line}} = \int_{\tilde{\tau}_{\mathrm{in}}}^{\tilde{\tau}_{\mathrm{out}}} g(\vec{x}(\tau))\,d\tau,
\label{eq:Iline_def}
\end{equation}
which admits a closed-form solution via the error function (Props.~3--4 in supp.\ mat.):
\begin{align}
I_{\mathrm{line}}
&=
\exp\!\Big(-\tfrac{1}{2}d_{\perp}^2\Big)\,
\sqrt{\frac{\pi}{2A}}\,
\Big[
\erf\!\Big(\sqrt{\tfrac{A}{2}}(\tilde{\tau}_{\mathrm{out}}-\tau_c)\Big)
-
\erf\!\Big(\sqrt{\tfrac{A}{2}}(\tilde{\tau}_{\mathrm{in}}-\tau_c)\Big)
\Big].
\label{eq:Iline_closed}
\end{align}

\subsection{Normalization by Maximum Line Integral}
\label{sec:normalization}
To make the integral dimensionless and comparable across Gaussians of varying size, we normalize by the maximum attainable line integral within the same $\chi^2$ bound: a center pass ($d_{\perp}^2=0$) through the longest axis $s_{\max}=\max(\gscale)$ (proof in Sec.~2.5 of supp.\ mat.),
\begin{equation}
I_{\max}
=
\sqrt{2\pi}\;\frac{s_{\max}}{\|\vec{v}\|_2}\;
\erf\!\Big(\sqrt{\tfrac{\chi^2}{2}}\Big),
\label{eq:Imax_global}
\end{equation}
where the factor $s_{\max}/\|\vec{v}\|_2$ converts between the segment parameter $\tau\in[0,1]$ of Eq.~\ref{eq:ray_seg} and world distance along the ray. We then define the normalized thickness factor
\begin{equation}
\eta = \clamp\!\Big(\frac{I_{\mathrm{line}}}{I_{\max}},\,0,\,1\Big).
\label{eq:w_factor}
\end{equation}
The factor $\eta\in[0,1]$ (Prop.~5 in supp.\ mat.) is near $0$ for grazing intersections and approaches $1$ for a center pass along the major axis.

\subsection{From Opacity to Shadow Transmittance}
\label{sec:transmittance}
Each occluder Gaussian carries an opacity parameter $\alpha\in(0,1)$.
We update the shadow transmittance $T$ multiplicatively over all Gaussians intersecting the ray (found via BVH traversal and the exact test above). The normalized thickness factor $\eta$ modulates each Gaussian's effective attenuation. We consider three update rules.

\paragraph{Linear normalized model.}
Using $\eta$ as a fractional coverage, we set
\begin{equation}
T \leftarrow T\,(1-\alpha\,\eta).
\label{eq:mode_linear}
\end{equation}

\paragraph{Exponentiated normalized model.}
A more thickness-consistent variant treats $\eta$ as a fraction of optical thickness,
\begin{equation}
T \leftarrow T\,(1-\alpha)^{\eta},
\label{eq:mode_power}
\end{equation}
which is equivalent to attenuating with an effective opacity (Prop.~6 in supp.\ mat.)
\begin{equation}
\alpha_{\mathrm{eff}} = 1-(1-\alpha)^{\eta},
\qquad
T \leftarrow T\,(1-\alpha_{\mathrm{eff}}).
\label{eq:alpha_eff}
\end{equation}

\paragraph{Beer--Lambert integral model.}
Mapping opacity to an extinction coefficient $\kappa(\alpha)=-\ln(1-\alpha)\ge0$ gives the physically motivated update (Prop.~7 in supp.\ mat.)
\begin{equation}
T \leftarrow T\,\exp\!\big(-\kappa(\alpha)\,I_{\mathrm{line}}\big).
\label{eq:beer_lambert}
\end{equation}
We use the exponentiated model in all experiments (ablation in Sec.~\ref{sec:ablations}).

\subsection{Acceleration via BVH over Conservative AABBs}
\label{sec:bvh}
To accelerate intersection queries, we build a BVH over per-Gaussian axis-aligned bounding boxes that conservatively enclose the $\chi^2$ ellipsoid in world space. The ellipsoid projection radius along world axis $d\in\{x,y,z\}$ is
\begin{equation}
e_d = \sqrt{\chi^2\;\big(s_x^2 \grot_{d1}^2+s_y^2 \grot_{d2}^2+s_z^2 \grot_{d3}^2\big)},
\label{eq:aabb_extent}
\end{equation}
yielding $\mathrm{AABB}=[\gmean-\vec{e},\;\gmean+\vec{e}]$.
Only BVH candidates invoke the exact quadratic test Eqs.~\ref{eq:quad}--\ref{eq:tin_tout} and integral Eq.~\ref{eq:Iline_closed}, enabling efficient soft shadow estimation in dense 3DGS scenes. We implement the BVH using NVIDIA OptiX, leveraging hardware RT cores for near real-time performance (Sec.~\ref{sec:timing}).

\subsection{Shadow Rendering}
\label{sec:shadow_rendering}
After computing the transmittance $T_k \in [0,1]$ for each scene Gaussian $k$, we modulate its color during 3DGS rasterization. The shadowed color of Gaussian $k$ is $\gcolor_k' = T_k \, \gcolor_k$, where $T_k = 1$ indicates full illumination and $T_k \to 0$ indicates full shadow. The modified colors $\gcolor_k'$ are then splatted using standard $\alpha$-blending, producing the final shadowed rendering.

\subsection{Avatar Proxy for Temporally Stable Shadows}
\label{sec:shadow_proxy}

\begin{wrapfigure}{r}{0.5\linewidth}
    \vspace{-8mm}
    \centering
    \includegraphics[width=\linewidth]{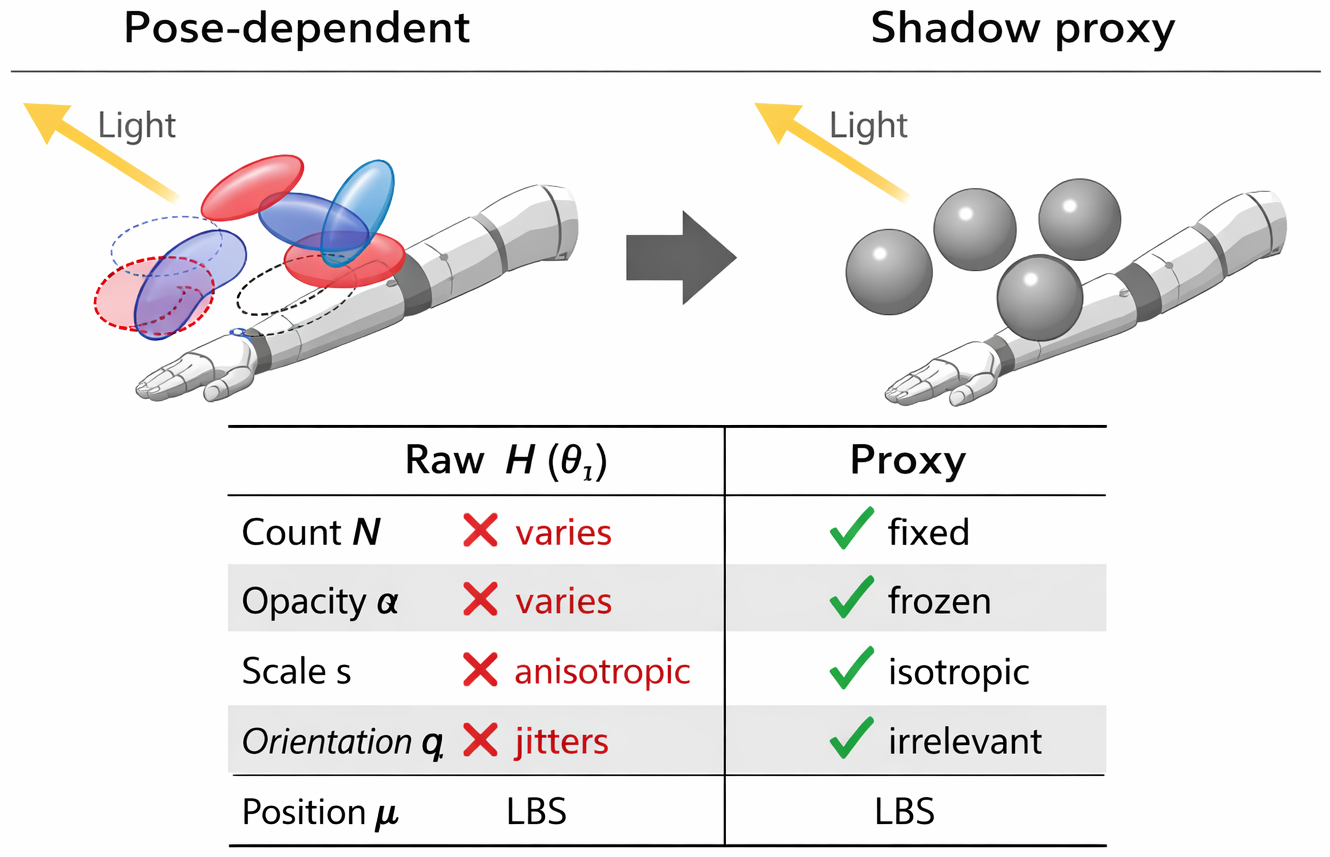}
    \caption{\textbf{Avatar proxy.} We replace the pose-dependent anisotropic Gaussians (left) with a stable isotropic proxy (right) that eliminates temporal flickering in shadows.}
    \label{fig:proxy}
    \vspace{-4mm}
\end{wrapfigure}
While the pose-dependent 3DGS avatar model produces high-fidelity renderings, using the raw primitives for shadow casting introduces temporal instability. The pose-to-geometry generator $\mathcal{H}(\pose_t)$, a neural network that maps body pose $\pose_t$ to 3D Gaussian primitives, produces outputs that fluctuate in topology, opacity, and anisotropic orientation to satisfy view-dependent photometric losses. When used as shadow occluders, these fluctuations cause the shadow silhouette to flicker and oscillate rapidly between frames.

To address this, we decouple the render-ready geometry from the shadow-casting geometry by constructing a \textbf{stable avatar proxy} (Fig.~\ref{fig:proxy}), a regularized derivative of the avatar that guarantees temporal coherence.

\paragraph{Canonical topology locking.}
To eliminate topological noise, we freeze the generative process at a canonical reference pose $\pose_{\mathrm{ref}}$ (e.g., a T-pose), caching the configuration once:
\begin{equation}
    \mathcal{P}_{\mathrm{proxy}} = \{ (\gmean_k^{(0)}, \gscale_k^{(0)}, \alpha_k^{(0)}) \}_{k=1}^N = \mathcal{H}(\pose_{\mathrm{ref}}).
\end{equation}
By fixing the number of primitives $N$ and their initial attributes, we ensure a persistent set of occluders, preventing artifacts caused by primitives appearing and disappearing between frames.

\paragraph{Isotropic regularization.}
To enforce rotational invariance, we override the anisotropic scale with a conservative isotropic radius $r_k = \max(s_{k,x}^{(0)}, s_{k,y}^{(0)}, s_{k,z}^{(0)})$, transforming the covariance into $\gcovar'_k = r_k^2 \mat{I}$. Since a sphere's silhouette is rotation-invariant, the shadow projection becomes independent of the primitive's orientation.

\paragraph{Simplified kinematics.}
During inference, we articulate the proxy using Linear Blend Skinning (LBS) applied strictly to positions. Given the canonical center $\gmean_k^{(0)}$ and joint transforms $\mat{M}_j(\pose_t)$:
\begin{equation}
    \gmean_k^{(t)} = \sum_{j \in \mathcal{J}} w_{kj} \mat{M}_j(\pose_t) \gmean_k^{(0)},
\end{equation}
where $\mathcal{J}$ denotes the skeleton joints and $w_{kj}\ge 0$ are normalized skinning weights ($\sum_{j\in\mathcal{J}}w_{kj}=1$). Crucially, we do \textbf{not} update the orientation of the proxies. Since the primitives are isotropic spheres, their rotation is irrelevant to the ray intersection test, guaranteeing that the shadow signal is continuous with respect to the pose $\pose_t$.

\section{Experiments}
\label{sec:experiments}

\definecolor{best}{RGB}{220,250,220}
\definecolor{second}{RGB}{240,245,220}
\definecolor{headerbg}{RGB}{200,220,240}

\begin{table}[t]
\fontsize{7}{8}\selectfont
\setlength{\tabcolsep}{2pt}
\renewcommand{\arraystretch}{1.1}
\begin{minipage}[t]{0.58\linewidth}
\centering
\caption{\textbf{Shadow quality vs.\ pseudo-GT on ScanNet++.} Averaged over 5 scenes. Mesh Shadow is an oracle.}
\label{tab:shadow-quality}
\begin{tabular}{@{}lccc@{}}
\toprule
\rowcolor{headerbg}
\textbf{Method} & \textbf{SAE}$\downarrow$ & \textbf{SM-IoU}$\uparrow$ & \textbf{BF}$\uparrow$ \\
\midrule
\rowcolor{second}
\textbf{Mesh Shadow (oracle)} & \textbf{0.000} & \textbf{1.000} & \textbf{1.000} \\
\rowcolor{best}
\textbf{Ours (RAGA)} & \textbf{0.031} & \textbf{0.847} & \textbf{0.812} \\
3DGRT~\cite{loccoz20243dgrt} (mod.) & 0.058 & 0.741 & 0.693 \\
RaySplat~\cite{Byrski2025RaySplats} (mod.) & 0.062 & 0.728 & 0.679 \\
\bottomrule
\end{tabular}
\end{minipage}
\hfill
\begin{minipage}[t]{0.38\linewidth}
\centering
\caption{\textbf{Temporal consistency (TSC$\downarrow$).} Averaged over 5 sequences. Lower = more stable.}
\label{tab:temporal}
\begin{tabular}{@{}lc@{}}
\toprule
\rowcolor{headerbg}
\textbf{Method} & \textbf{TSC}$\downarrow$ \\
\midrule
\rowcolor{best}
\textbf{Ours (full proxy)} & \textbf{0.0018} \\
Ours w/o proxy & 0.0061 \\
3DGRT~\cite{loccoz20243dgrt} (mod.) & 0.0114 \\
RaySplat~\cite{Byrski2025RaySplats} (mod.) & 0.0121 \\
\bottomrule
\end{tabular}
\end{minipage}
\vspace{-2mm}
\end{table}

\newcommand{\compw}{0.245\linewidth}
\begin{figure*}[t]
    \centering
    \setlength{\tabcolsep}{1pt}
    \renewcommand{\arraystretch}{0.6}
    \begin{tabular}{cccc}
        {\small RaySplat~\cite{Byrski2025RaySplats} (mod.)} &
        {\small 3DGRT~\cite{loccoz20243dgrt} (mod.)} &
        {\small Ours (RAGA)} &
        {\small Pseudo-GT} \\[2pt]
        \includegraphics[width=\compw]{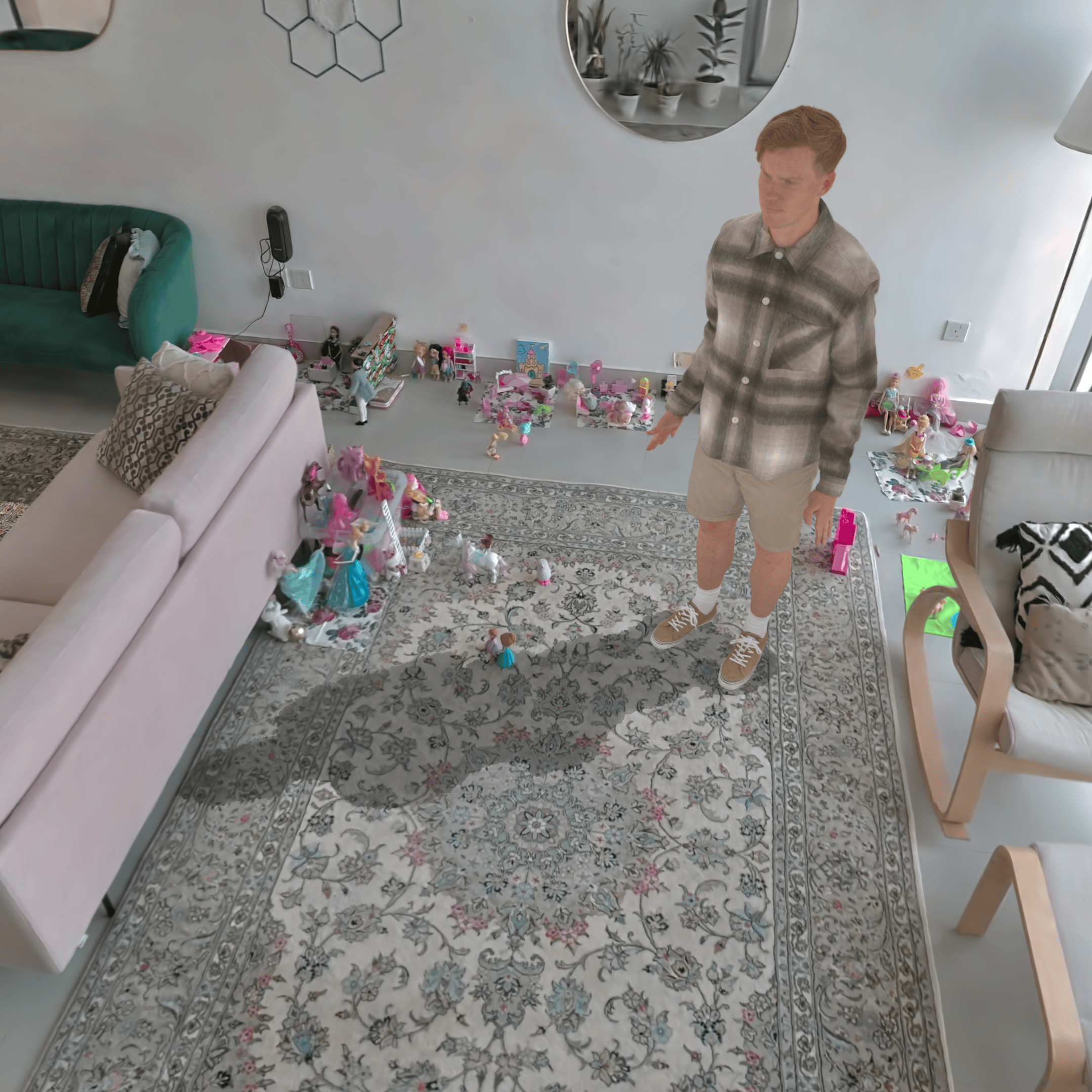} &
        \includegraphics[width=\compw]{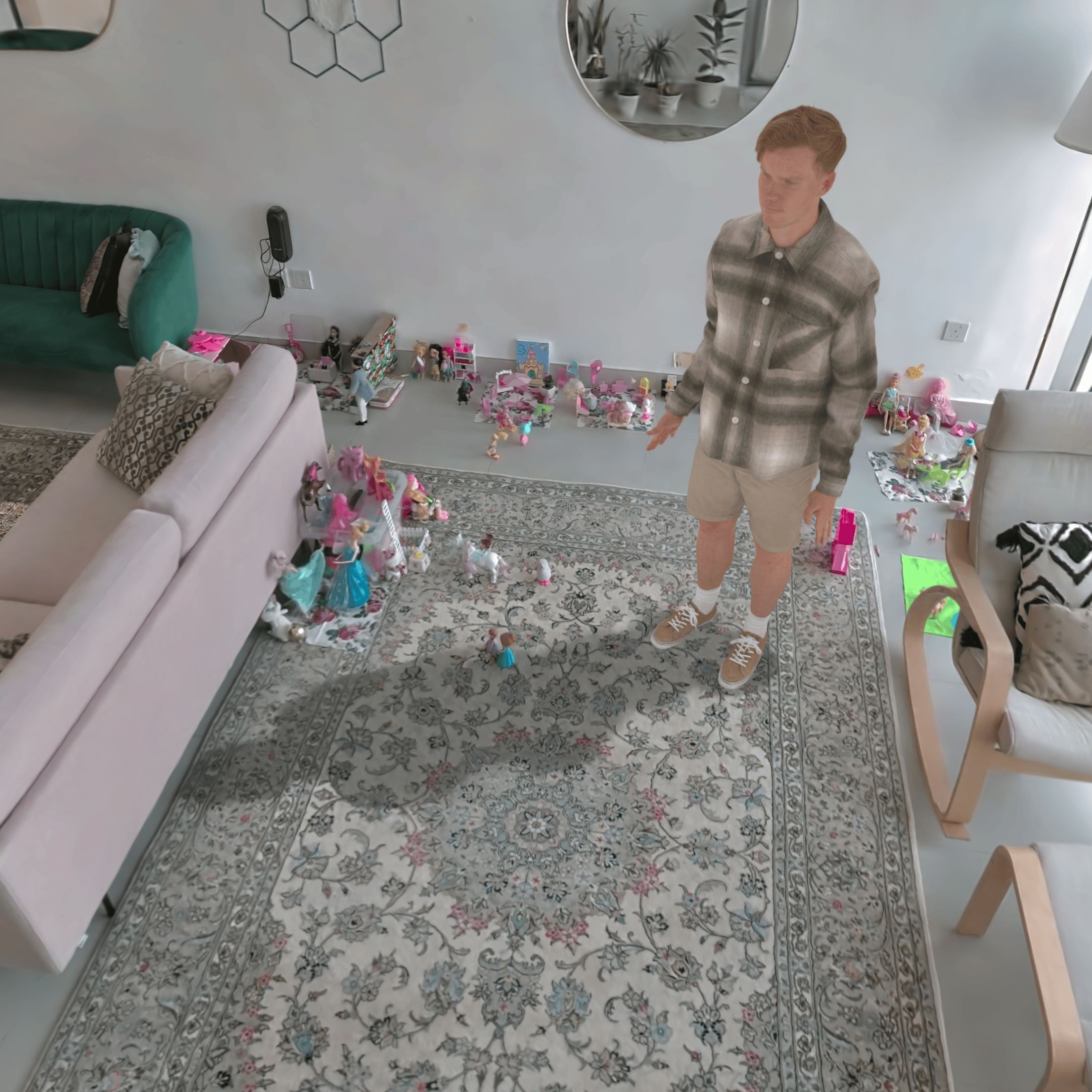} &
        \includegraphics[width=\compw]{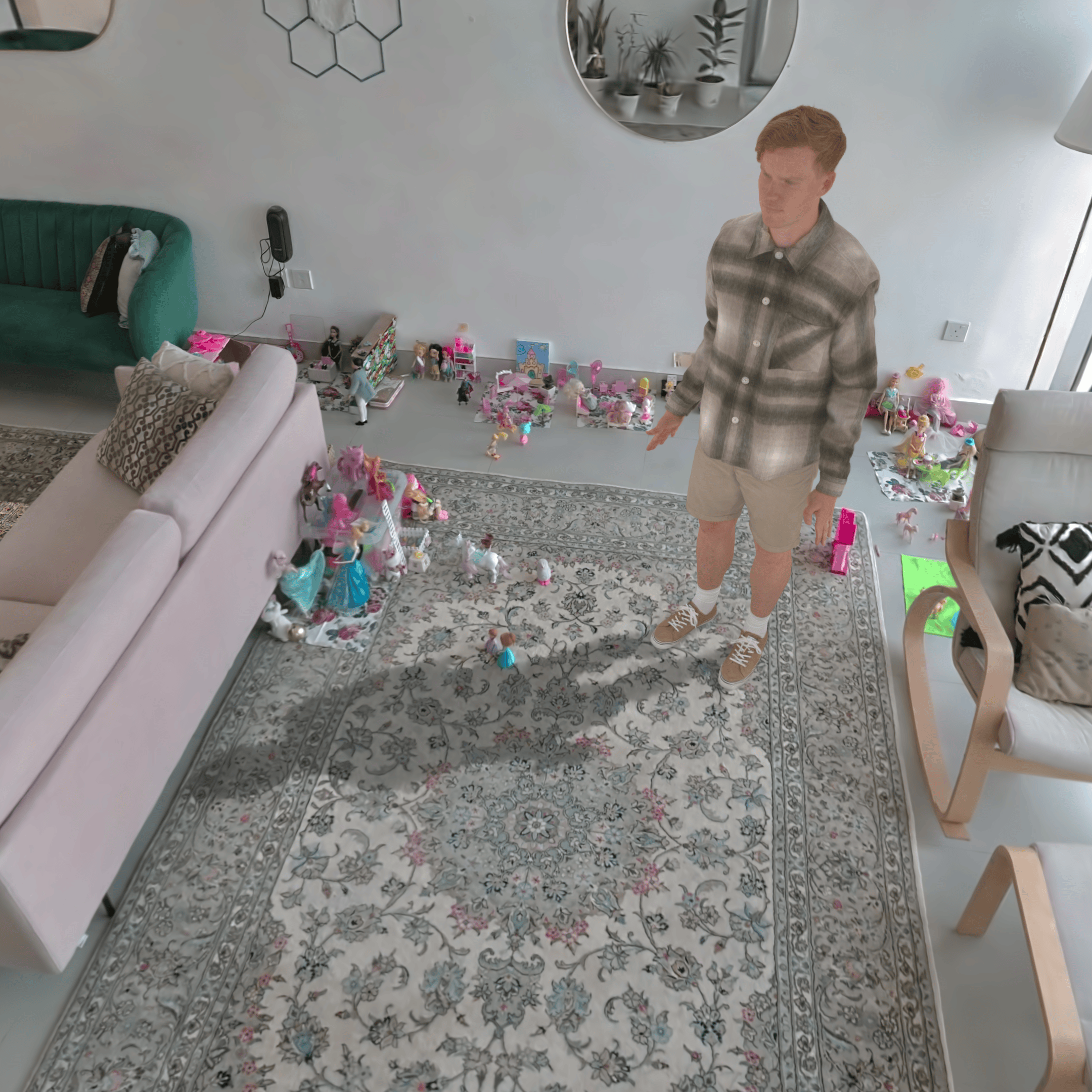} &
        \includegraphics[width=\compw]{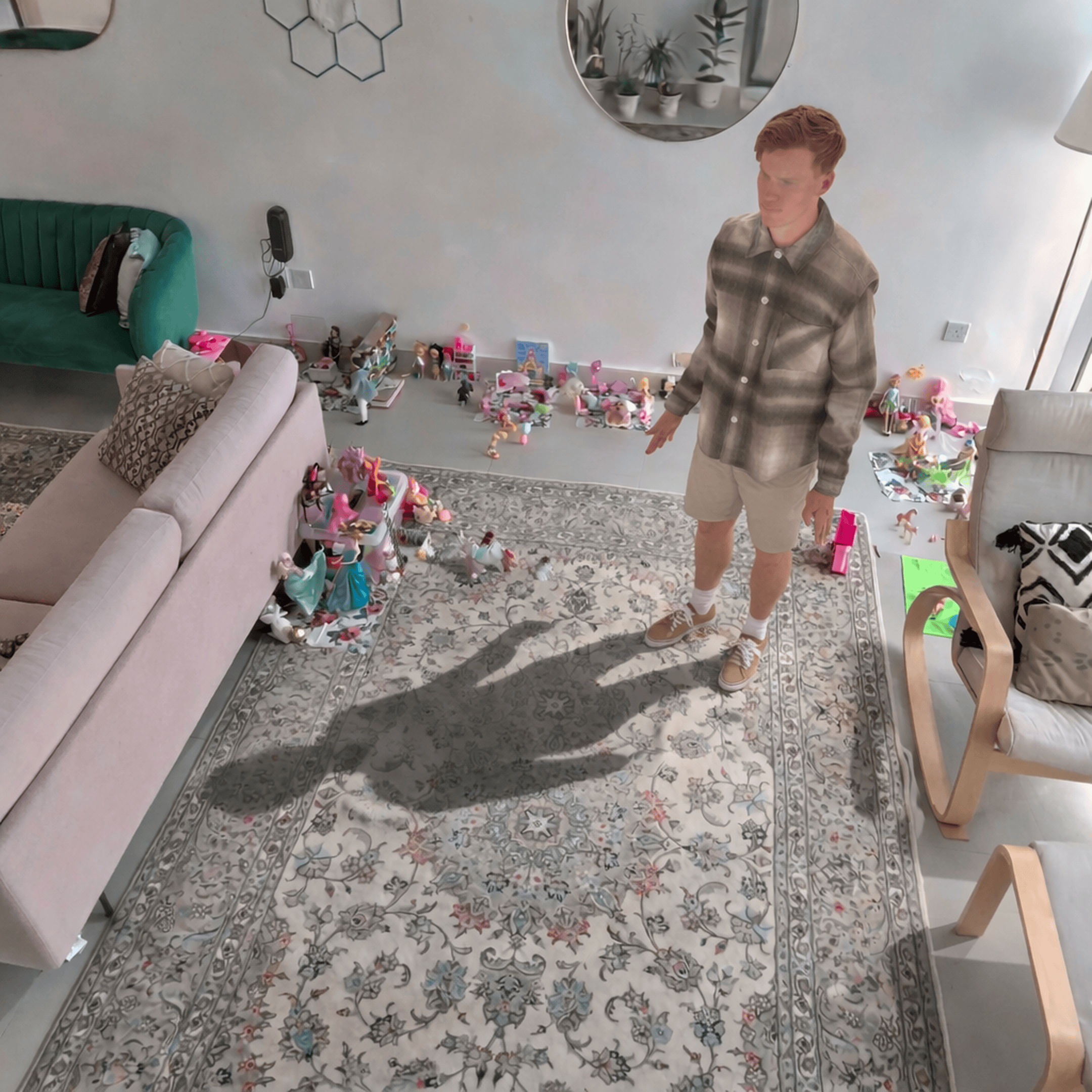} \\[1pt]
        \includegraphics[width=\compw]{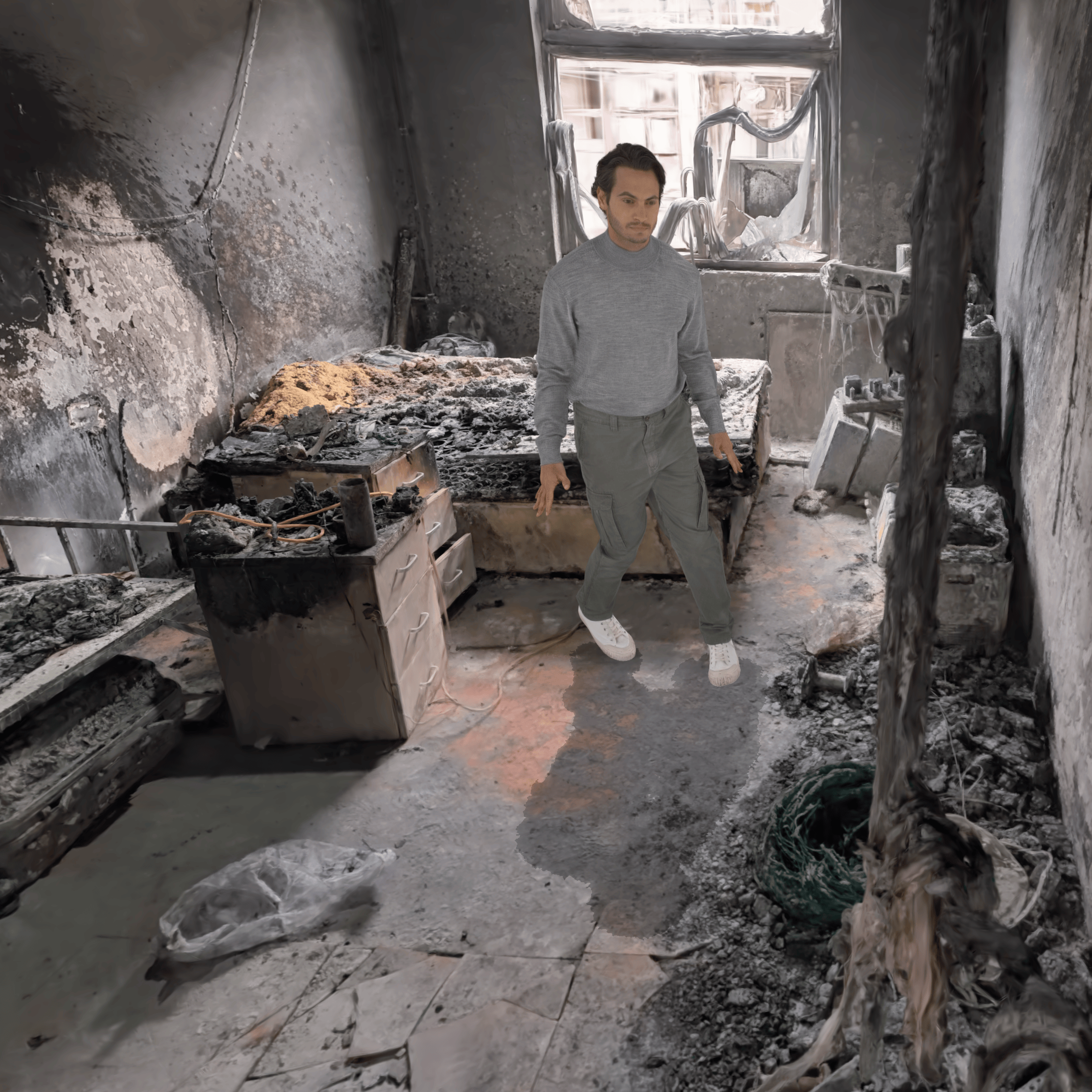} &
        \includegraphics[width=\compw]{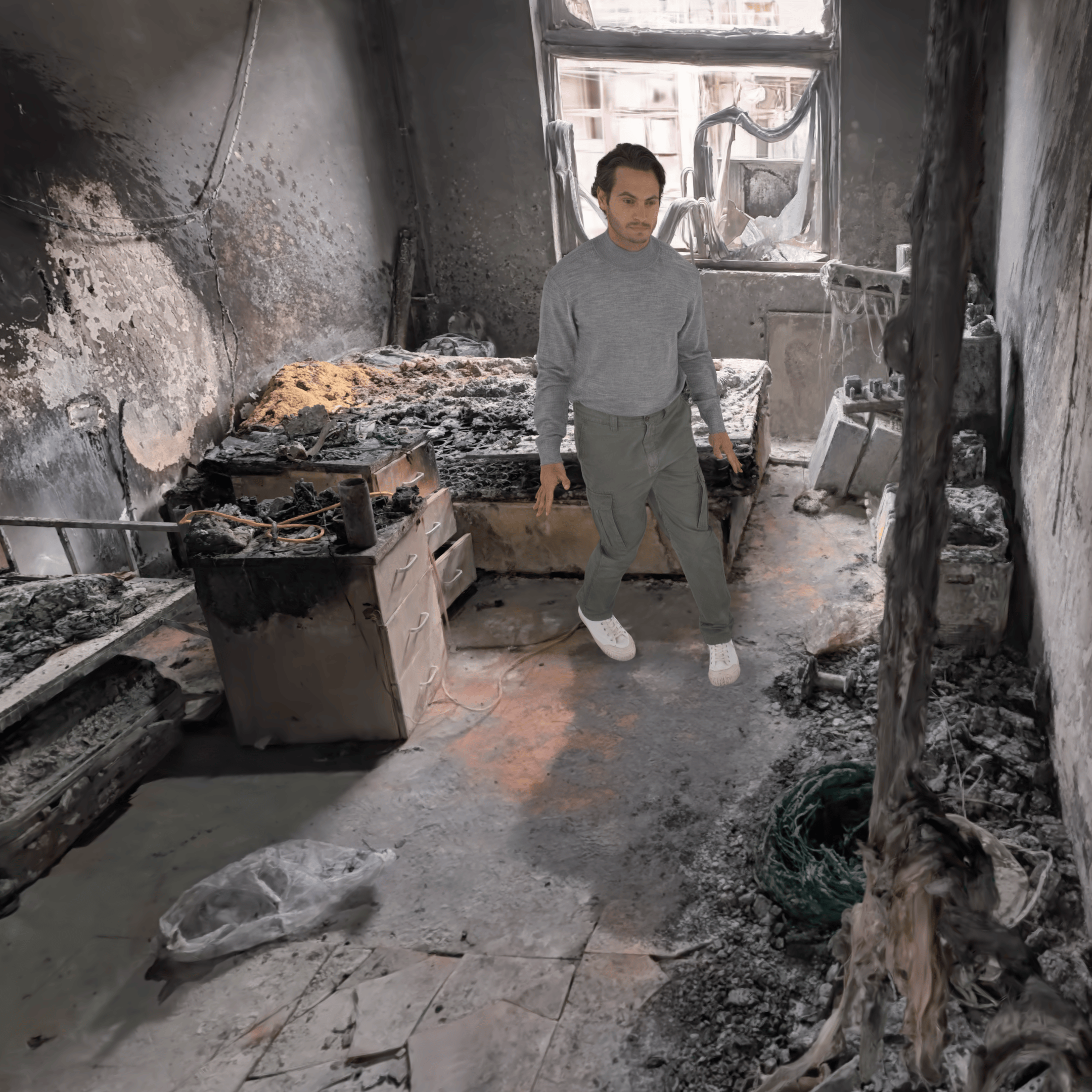} &
        \includegraphics[width=\compw]{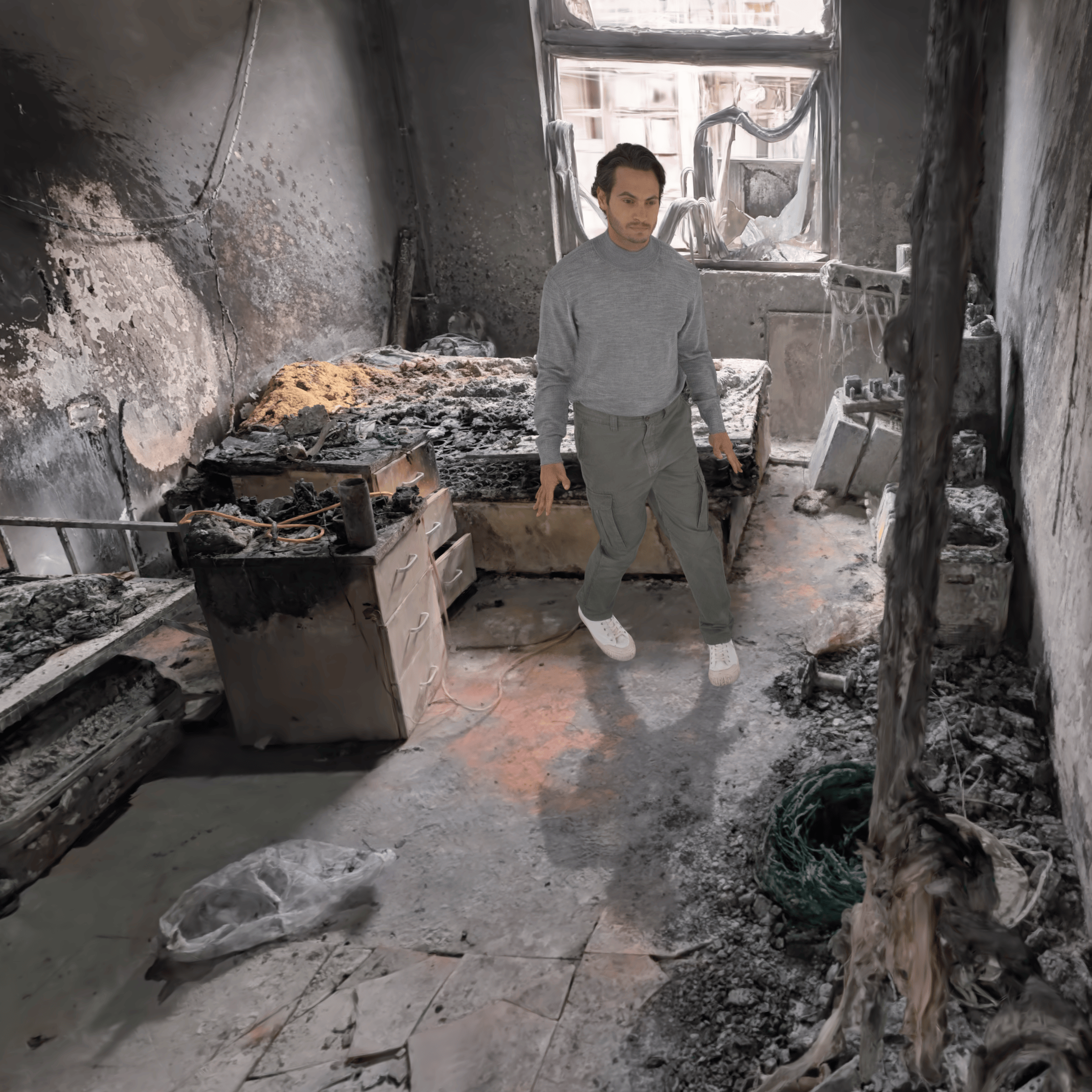} &
        \includegraphics[width=\compw]{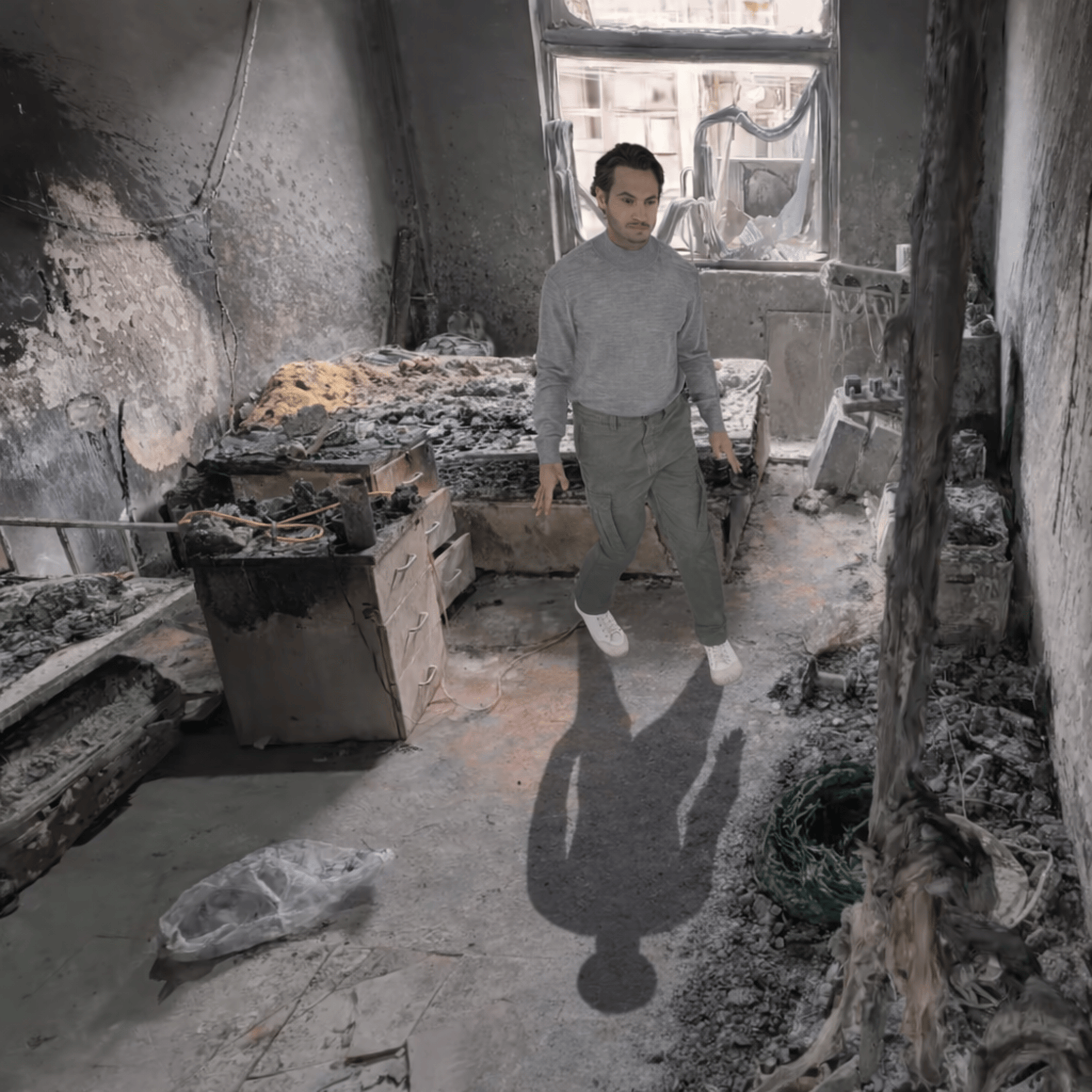} \\ [1pt]
    \end{tabular}
    \caption{Qualitative comparison against baselines on SuperSplat~\cite{supersplat} scenes. Pseudo-GT is obtained using a proxy floor mesh and posed SMPL mesh via classical shadow mapping. Our method produces soft, physically plausible shadows that closely match the pseudo-GT. Both 3DGRT~\cite{loccoz20243dgrt} and RaySplat~\cite{Byrski2025RaySplats} lose shadow detail due to their implausible attenuation models.}
    \label{fig:comparison}
\vspace{-4mm}
\end{figure*}

We evaluate RAGA along two axes: \emph{shadow quality}, how accurately do our ray-traced Gaussian shadows approximate mesh-based ground truth? (Sec.~\ref{sec:shadow-quality}); and \emph{temporal stability}, does the avatar proxy reduce flickering compared to raw Gaussian shadow casting? (Sec.~\ref{sec:temporal}); in both evaluations we compare with two modified ray-traced methods, 3DGRT~\cite{loccoz20243dgrt} and RaySplat~\cite{Byrski2025RaySplats}. We further report \emph{ablations} on each pipeline component (Sec.~\ref{sec:ablations}) and a timing analysis (Sec.~\ref{sec:timing}).

\subsection{Experimental Setup}
\label{sec:setup}

\textbf{Pseudo-GT Protocol.}
We obtain pseudo-GT shadow maps $S^\dagger(p)$ via classical mesh-based rendering in two settings:
\begin{itemize}
    \item \textbf{ScanNet++~\cite{yeshwanthliu2023scannetpp} scenes:} Since ScanNet++ provides both a 3DGS and a mesh reconstruction, we replace the avatar with the posed SMPL mesh and cast shadows onto the scene mesh using classical shadow mapping. We render the shadow attenuation onto a transparent surface and composite it with the 3DGS scene rendering. We use 5 ScanNet++ scenes for quantitative evaluation.
    \item \textbf{General 3DGS scenes (e.g.\ SuperSplat~\cite{supersplat}):} For scenes without a mesh reconstruction, we construct an approximate floor plane from the 3DGS point cloud, replace the avatar with the posed SMPL mesh, and cast shadows onto the floor proxy via classical rendering. The floor is set to invisible and the resulting shadow map is composited with the 3DGS rendering.
\end{itemize}

\paragraph{Why not use mesh-based shadow casting?}
While the pseudo-GT protocol above uses meshes to establish a reference, mesh-based shadows are not a practical solution for general 3DGS scenes:
\begin{itemize}
    \item \textbf{Scene mesh extraction is lossy.} Obtaining a high-quality mesh from a 3DGS scene is not straightforward and leads to significant loss of detail~\cite{ye2024gaustudio}. While we use a floor proxy for pseudo-GT, this discards all wall- and furniture-based shadow receivers; ScanNet++ is an exception where dense meshes happen to be available.
    \item \textbf{Mesh proxies lose detail.} Mesh-based shadow casting requires replacing occluders with mesh proxies such as SMPL for humans, which discards clothing and hair detail from the shadow silhouette. While our avatar proxy also smooths fine detail, it retains significantly more geometric fidelity than the bare SMPL mesh because it operates on the actual Gaussian point cloud rather than a parametric body model.
    \item \textbf{No support for Gaussian objects.} Mesh proxies do not generalize to arbitrary 3DGS objects inserted into scenes. Our method operates entirely in Gaussian space and naturally supports avatar--object interactions, as shown in the qualitative results (Sec.~\ref{sec:qual}).
\end{itemize}

\textbf{Avatars.}
We use 3DGS avatars from ActorsHQ~\cite{isik2023humanrf} (48 cameras) and AvatarReX~\cite{zheng2023avatarrex}, spanning a range of reconstruction quality. For multi-avatar and avatar-object scenarios, we additionally use NeuralDome~\cite{zhang2023neuraldome}.

\textbf{Baselines.}
Since no prior method addresses the same problem (shadow casting for animated 3DGS avatars in 3DGS scenes), we construct the strongest possible Gaussian-based baselines by modifying two recent ray-tracing methods for shadow casting:
\begin{itemize}
    \item \textbf{3DGRT}~\cite{loccoz20243dgrt} (modified): Following 3DGRT, we first detect ray--Gaussian intersections using an icosahedron approximation around each Gaussian, then for detected intersections we accumulate opacity values to compute ray attenuation. This formulation does not use our line integral or avatar proxy.
    \item \textbf{RaySplat}~\cite{Byrski2025RaySplats} (modified): We detect ray--Gaussian intersections using the exact analytic test from RaySplat, then accumulate opacity for each detected Gaussian. As with 3DGRT, this formulation uses neither our line integral nor avatar proxy.
    \item \textbf{Mesh Shadow (oracle)}: We render a shadow-only pass by casting shadows from the posed SMPL mesh onto a transparent mesh receiver via classical shadow mapping, then alpha-blend the result onto the 3DGS scene rendering.
\end{itemize}

\subsection{Shadow Quality Evaluation}
\label{sec:shadow-quality}

We evaluate shadow maps against the mesh-based pseudo-GT using three pixel-space metrics computed in an avatar-centric region of interest:
\begin{itemize}
    \item \textbf{SAE} (Shadow Attenuation Error): Mean absolute error between predicted and pseudo-GT shadow within the shadow region $\Omega_s$ (lower is better).
    \item \textbf{SM-IoU} (Shadow Matte IoU): Intersection-over-union of binarized shadow masks (threshold $0.1$) (higher is better).
    \item \textbf{BF} (Boundary F-measure): F-score between shadow boundaries with 2\,px tolerance (higher is better).
\end{itemize}

Tab.~\ref{tab:shadow-quality} reports shadow quality averaged over 5 ScanNet++ scenes. Our method achieves the best results among all Gaussian-based approaches, approaching the mesh oracle. 3DGRT's binary hit test based on its icosahedron approximation produces hard shadow boundaries without penumbrae (Fig.~\ref{fig:method}\,\textcolor{figA}{a}), leading to poor SAE and BF scores. RaySplat's collision-based approach reduces to a max-response evaluation (Fig.~\ref{fig:method}\,\textcolor{figC}{c}--\textcolor{figE}{e}) that lacks proper attenuation modeling. Fig.~\ref{fig:comparison} shows a visual comparison.

\begin{table*}[t]
\fontsize{6.5}{7.5}\selectfont
\setlength{\tabcolsep}{2pt}
\renewcommand{\arraystretch}{1.1}
\begin{minipage}[t]{0.62\linewidth}
\centering
\caption{\textbf{Ablation studies.} Evaluated against pseudo-GT on ScanNet++. Best in each group is \textbf{bold}.}
\label{tab:ablations}
\begin{tabular}{@{}l|ccc|cc@{}}
\toprule
\rowcolor{headerbg}
 & \multicolumn{3}{c|}{\textit{(A) Attenuation model}} & \multicolumn{2}{c@{}}{\textit{(B) Normalization}} \\
\rowcolor{headerbg}
\textbf{Metric}
 & \rotatebox{45}{\textbf{Exp.\ (Ours)}}
 & \rotatebox{45}{Linear}
 & \rotatebox{45}{Beer--L.}
 & \rotatebox{45}{\textbf{$I_{\max}$ (Ours)}}
 & \rotatebox{45}{w/o norm.} \\
\midrule
\textbf{SAE} $\downarrow$ & \cellcolor{best}\textbf{0.031} & 0.035 & 0.039 & \cellcolor{best}\textbf{0.031} & 0.047 \\
\textbf{SM-IoU} $\uparrow$ & \cellcolor{best}\textbf{0.847} & 0.831 & 0.819 & \cellcolor{best}\textbf{0.847} & 0.793 \\
\textbf{BF} $\uparrow$ & \cellcolor{best}\textbf{0.812} & 0.796 & 0.781 & \cellcolor{best}\textbf{0.812} & 0.754 \\
\textbf{TSC} $\downarrow$ & \cellcolor{best}\textbf{0.0018} & 0.0021 & 0.0024 & \cellcolor{best}\textbf{0.0018} & 0.0022 \\
\bottomrule
\end{tabular}
\end{minipage}
\hfill
\begin{minipage}[t]{0.35\linewidth}
\centering
\caption{\textbf{Avatar proxy ablation.} Temporal stability (TSC$\downarrow$). Lower = more stable.}
\label{tab:proxy}
\begin{tabular}{@{}cc@{}}
\toprule
\rowcolor{headerbg}
\rotatebox{45}{\textbf{Variant}} & \rotatebox{45}{\textbf{TSC}$\downarrow$} \\
\midrule
\rowcolor{best}
\textbf{Full proxy (Ours)} & \textbf{0.0018} \\
w/o topology lock & 0.0037 \\
w/o isotropic reg. & 0.0032 \\
No proxy & 0.0061 \\
\bottomrule
\end{tabular}
\end{minipage}
\vspace{-2mm}
\end{table*}

\newcommand{\ablw}{0.32\linewidth}
\begin{figure}[t]
    \centering
    \setlength{\tabcolsep}{2pt}
    \renewcommand{\arraystretch}{0.6}
    \begin{tabular}{ccc}
        {\scriptsize w/o $I_{\max}$ norm.} &
        {\scriptsize Linear atten.} &
        {\scriptsize Ours (full)} \\[2pt]
        \includegraphics[width=\ablw]{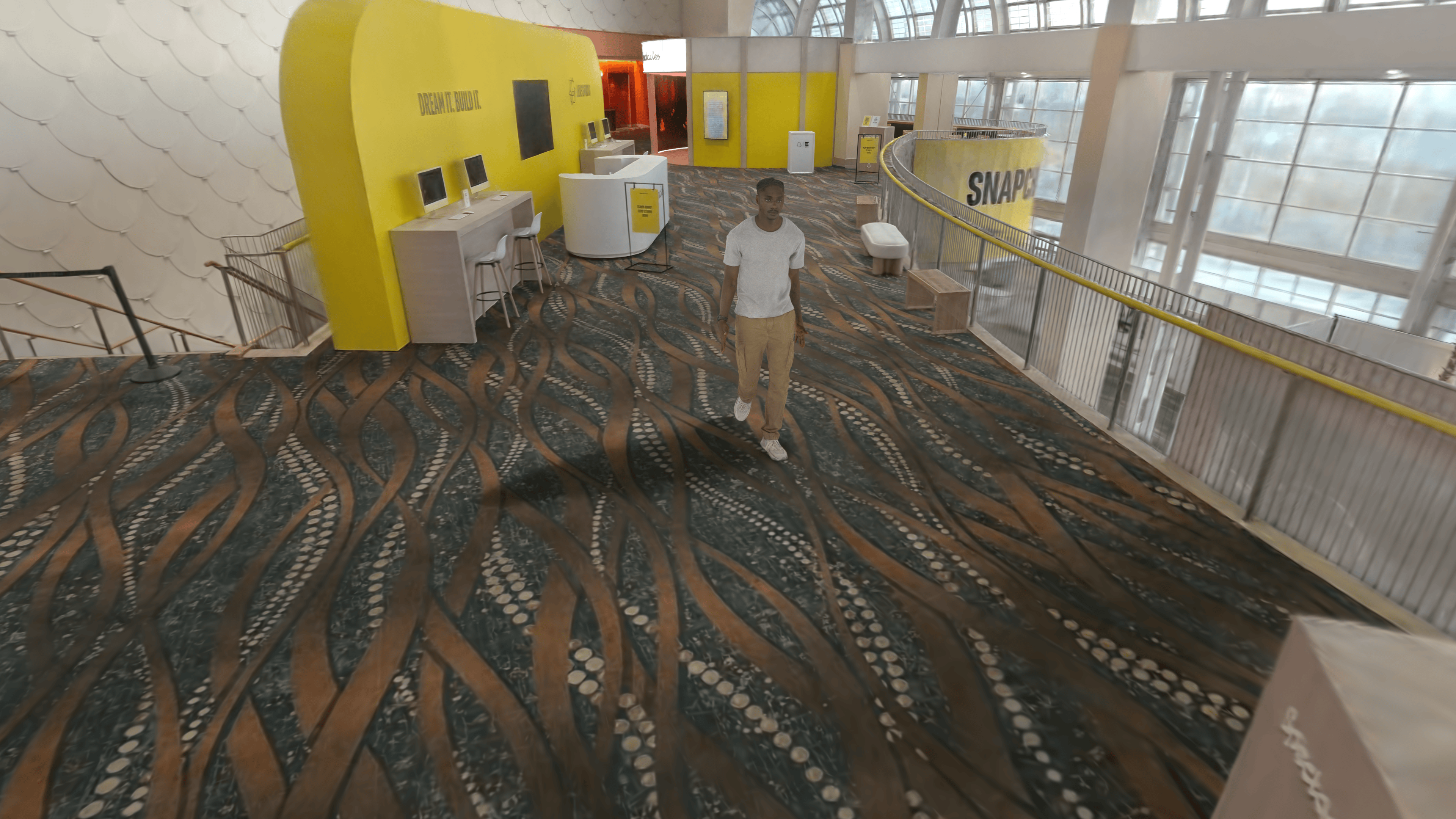} &
        \includegraphics[width=\ablw]{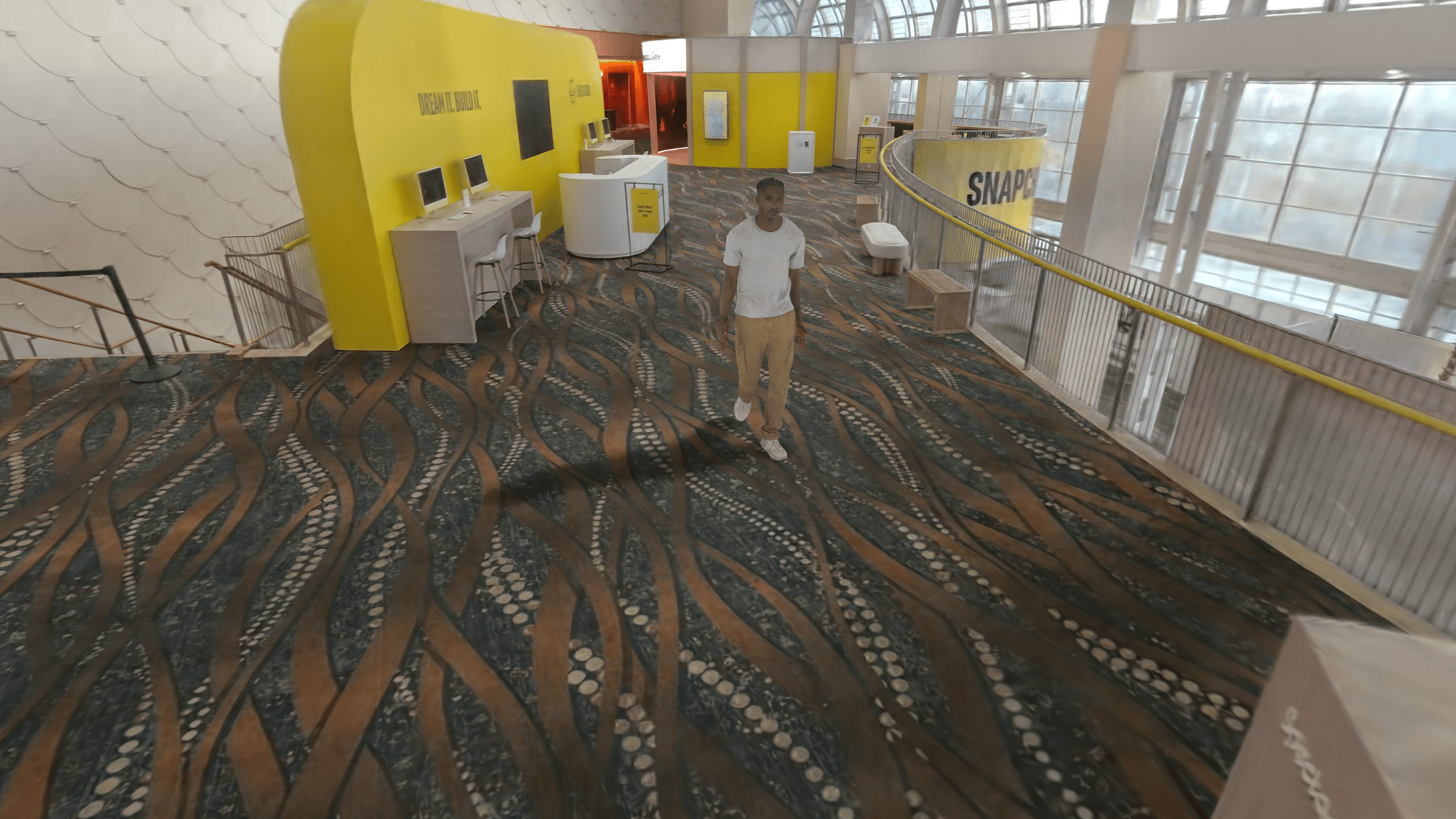} &
        \includegraphics[width=\ablw]{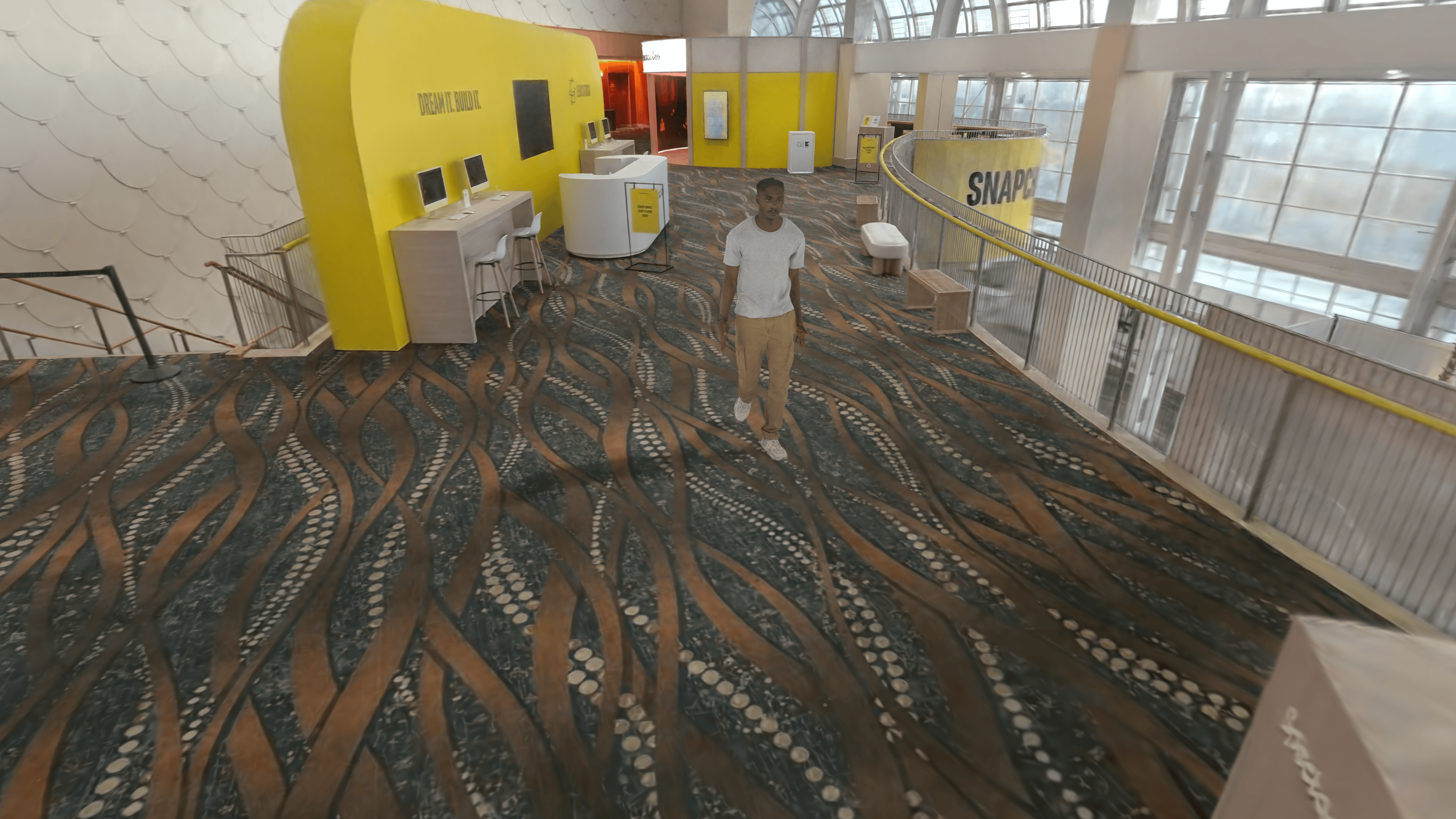} \\[1pt]
        {\scriptsize\color{red} Dark blobs} &
        {\scriptsize\color{red} Hard falloff} &
        {\scriptsize\color{green!50!black} Natural} \\
    \end{tabular}
    \caption{\textbf{Visual ablations.} Without $I_{\max}$ normalization (left), large anisotropic Gaussians cast disproportionately dark shadows. The linear attenuation model (middle) produces harder shadow falloff than our full exponentiated model (right).}
    \label{fig:ablation}
\vspace{-2mm}
\end{figure}

\subsection{Temporal Stability}
\label{sec:temporal}

\begin{figure}[t]
    \centering
    \begin{minipage}[t]{0.30\linewidth}
        \vspace{0pt}
        \centering
        \includegraphics[width=\linewidth,height=2.8cm]{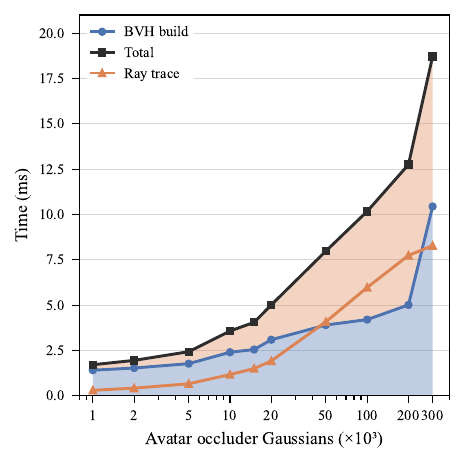}
        \caption{\textbf{Runtime} vs.\ avatar Gaussians.}
        \label{fig:runtime}
    \end{minipage}\hfill
    \begin{minipage}[t]{0.66\linewidth}
        \vspace{0pt}
        \centering
        \setlength{\tabcolsep}{1pt}
        \begin{tabular}{cc}
            \includegraphics[height=2.8cm]{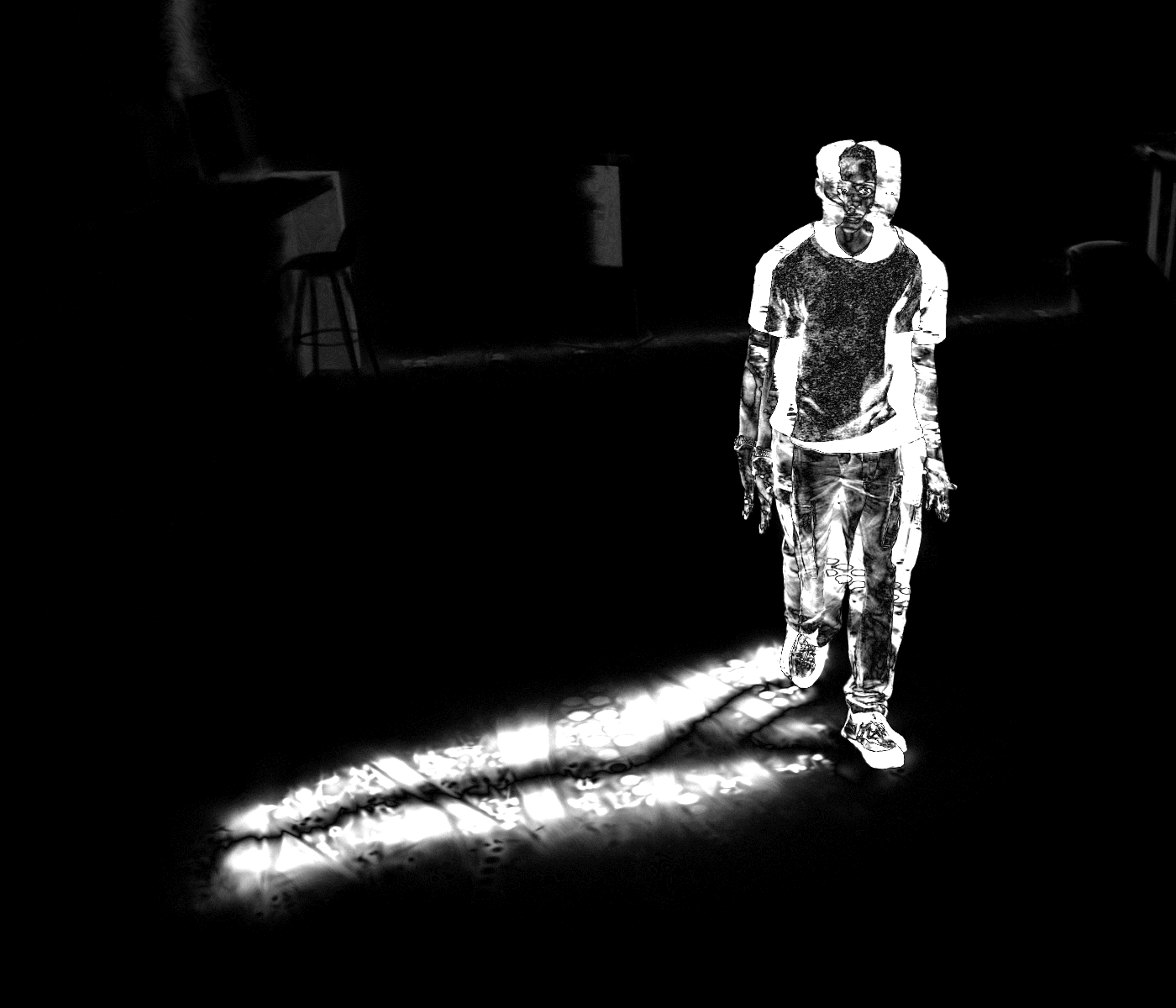} &
            \includegraphics[height=2.8cm]{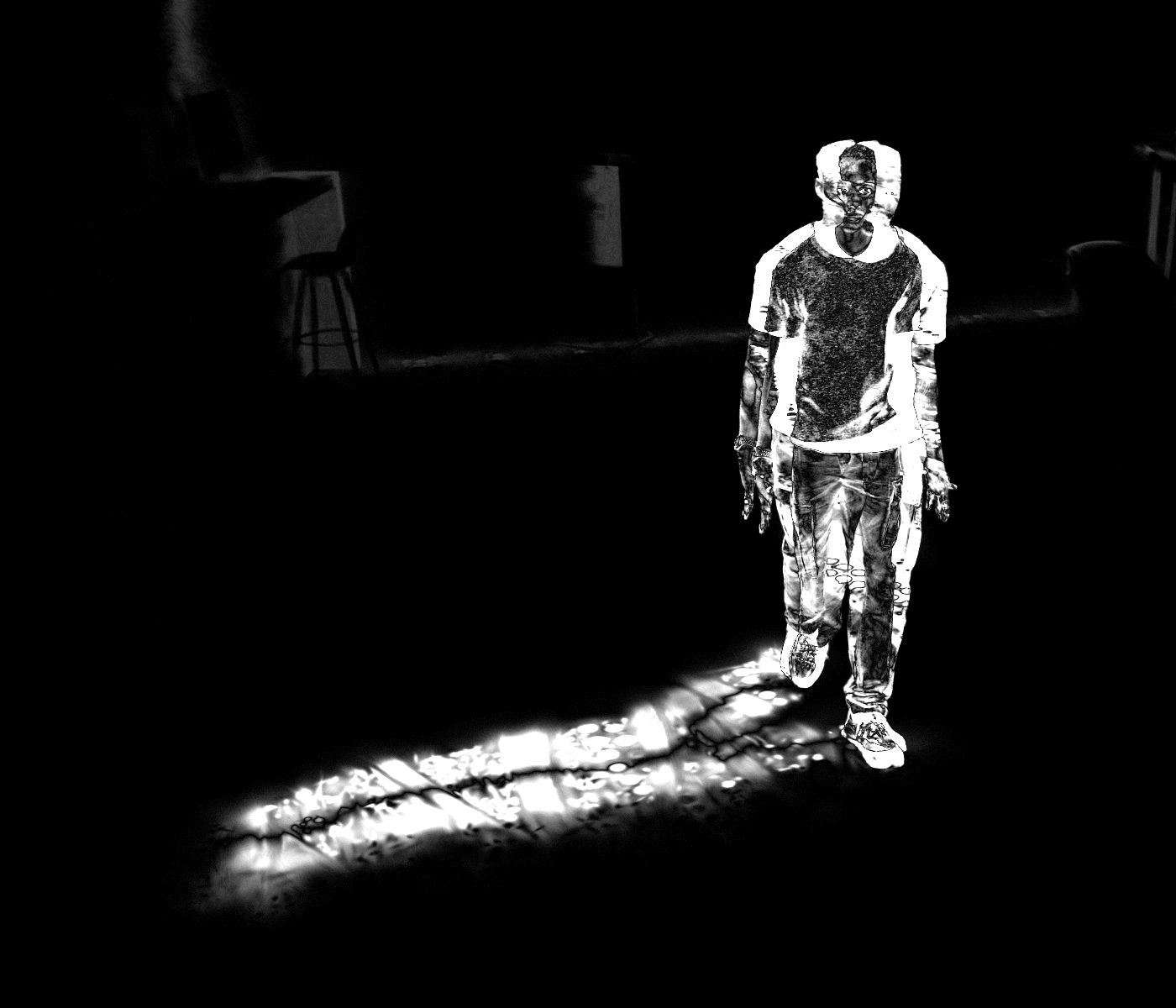} \\[1pt]
            {\scriptsize No proxy} & {\scriptsize Ours (proxy)} \\
        \end{tabular}
        \caption{\textbf{Proxy ablation.} $|S_t{-}S_{t-1}|$ maps: bright = large change. The proxy stabilizes shadows.}
        \label{fig:proxy_temporal}
    \end{minipage}
\vspace{-5mm}
\end{figure}

A key contribution of RAGA is the avatar proxy (Sec.~\ref{sec:shadow_proxy}) that ensures temporally stable shadows under animation. We measure temporal shadow consistency (TSC) as the mean per-pixel absolute difference between consecutive shadow frames: $\mathrm{TSC} = \frac{1}{F{-}1}\sum_{t=1}^{F-1} \frac{1}{|\Omega|}\sum_{p\in\Omega}|S_t(p) - S_{t-1}(p)|$, where $S_t$ is the shadow map at frame $t$, $F$ is the total number of frames, and $\Omega$ is the avatar-centric ROI. Lower TSC indicates more stable shadows. Tab.~\ref{tab:temporal} shows that the avatar proxy substantially reduces temporal flickering. Fig.~\ref{fig:proxy_temporal} shows temporal difference maps ($|S_t{-}S_{t-1}|$) confirming that the avatar proxy reduces shadow flickering.

\subsection{Perceptual Study}
\label{sec:perceptual}

We conduct a two-alternative forced-choice (2AFC) study with 12 naive raters across 5 ScanNet++ scenes, comparing \textbf{Ours (RAGA)} against a \textbf{No Shadow} baseline. Raters are shown 5 video clips (10\,s at 30\,fps) per scene with randomized method order and placement, and asked which rendering looks more realistic. On average, raters prefer RAGA 73\% of the time.

\subsection{Ablation Studies}
\label{sec:ablations}

We ablate each component of RAGA in Tab.~\ref{tab:ablations}. For the attenuation model, we compare linear, exponentiated, and Beer--Lambert transmittance updates (Eqs.~\ref{eq:mode_linear}--\ref{eq:beer_lambert}); the exponentiated model provides the best trade-off. For line integral normalization, we compare the raw line integral $I_{\mathrm{line}}$ against our normalized thickness factor $\eta = I_{\mathrm{line}}/I_{\max}$ (Eq.~\ref{eq:w_factor}); without normalization, large anisotropic Gaussians cast disproportionately dark shadows (Fig.~\ref{fig:ablation}). For the avatar proxy, we ablate canonical topology locking and isotropic regularization (Tab.~\ref{tab:proxy}).

\subsection{Runtime Analysis}
\label{sec:timing}

We analyze the runtime of RAGA as a function of avatar complexity on an NVIDIA H100 GPU with ${\sim}500$K scene Gaussians. Fig.~\ref{fig:runtime} reports per-frame shadow computation time as the number of avatar Gaussians varies from 20K to 400K. Only our shadow tracer, implemented as a CUDA kernel, runs at interactive rates of about 50\,FPS; the end-to-end pipeline (including 3DGS rasterization) achieves near real-time performance even for dense avatars thanks to OptiX hardware ray tracing and BVH acceleration. 

\subsection{Qualitative Results}
\label{sec:qual}

\begin{figure*}[t]
    \centering
    \includegraphics[width=0.99\linewidth]{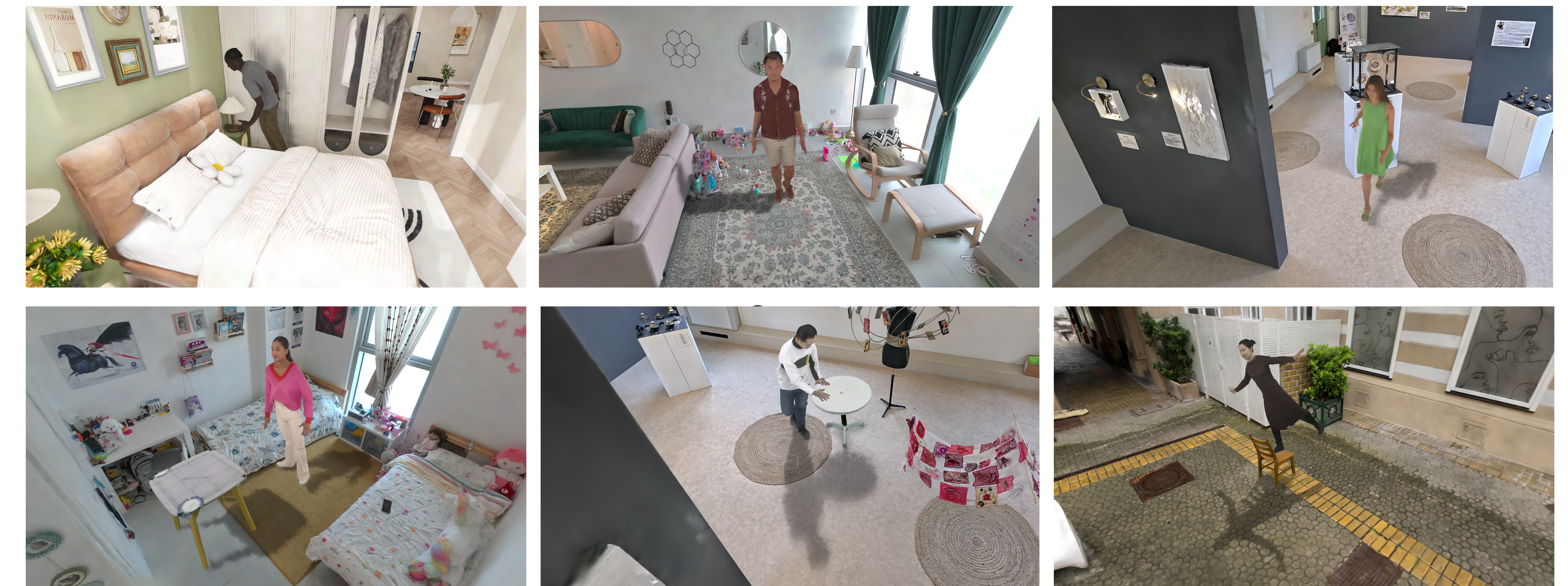}
    \caption{\textbf{Qualitative results}: our method casts plausible soft shadows for various 3DGS avatars animated alone and with objects across diverse scenes.}
    \label{fig:qualitative}
\vspace{-5mm}
\end{figure*}

Fig.~\ref{fig:qualitative} shows results across diverse scenarios: single-avatar animation, multi-avatar motion, and avatar-object interaction in various ScanNet++ and SuperSplat~\cite{supersplat} scenes. Our method produces soft, physically plausible shadows that follow the avatar's motion and respect scene geometry. We demonstrate results for multiple avatars interacting with scenes or avatars interacting with 3DGS objects placed in 3DGS scenes.
\vspace{-3mm}
\section{Conclusion}
\vspace{-2mm}

We presented RAGA, a principled and efficient framework for physically plausible shadow casting in animated 3DGS avatar-scene interaction scenarios. By operating entirely in Gaussian space, our approach overcomes a fundamental limitation of prior work, which relies on mesh-based shadow computation or approximate ray–Gaussian interactions that break down in dynamic settings.

At the core of RAGA is a ray-traced Gaussian shadow casting formulation that models light attenuation as a set of independent absorption events along rays cast toward light sources. Leveraging the closed-form solution of the ray–Gaussian line integral and analytically computable entry and exit points, our method enables stable and physically grounded aggregation of shadow contributions from multiple volumetric occluders. This principled formulation avoids the approximations commonly used in existing ray tracing approaches for 3DGS and results in significantly improved temporal stability and visual coherence.

To further address shadow instability in animated avatars, we introduced an avatar proxy that reduces variance caused by non-rigid deformations, leading to more consistent shadows during motion without sacrificing visual quality. Our GPU implementation integrates custom CUDA kernels with NVIDIA OptiX, exploiting hardware ray tracing to achieve near real-time performance. Extensive experiments across single-avatar, multi-avatar, and avatar–object interaction scenarios demonstrate that RAGA produces realistic, stable shadows.

While our method advances shadow casting for 3DGS avatars, several limitations remain. We model point and directional light but do not support area lights. Lighting on the avatar and scene remains baked into the 3DGS representation; we modulate scene Gaussian colors to create shadows but do not relight the avatar or scene under novel illumination. Similarly, we do not estimate light source positions from the scene; the user must specify them manually. Future work includes automatic light source estimation from 3DGS scenes, fully relightable avatars that respond to dynamic illumination, and area light support.
\vspace{-2mm}
\paragraph{\textbf{Acknowledgements.}}
We gratefully acknowledge support from the hessian.AI Service Center (funded by the Federal Ministry of Research, Technology and Space, BMFTR, grant no. 16IS22091) and the hessian.AI Innovation Lab (funded by the Hessian Ministry for Digital Strategy and Innovation, grant no. S-DIW04/0013/003). We especially thank Patrick Blauth at the hessian.AI Service Center for technical assistance. This work was further supported by the German Federal Ministry of Education and Research (BMBF): T\"ubingen AI Center, FKZ: 01IS18039A. GPM is a member of the Machine Learning Cluster of Excellence, EXC number 2064/1 -- Project number 390727645. The work was also partially supported by funding from KAUST -- Center of Excellence for Generative AI, under award number 5940.

\bibliographystyle{splncs04}
\bibliography{main}
\end{document}